\documentclass[10pt,twocolumn,letterpaper]{article}

\usepackage[pagenumbers]{cvpr} 

\usepackage{graphicx}
\usepackage{amsmath}
\usepackage{amssymb}
\usepackage{booktabs}
\usepackage{color,colortbl,xcolor}
\usepackage{caption}
\usepackage{subcaption}
\usepackage{physics}
\usepackage{multirow}
\usepackage{threeparttable}
\usepackage{subfloat}
\usepackage{tabularx}
\usepackage{adjustbox}
\usepackage{makecell}
\usepackage[accsupp]{axessibility}  
\usepackage[pagebackref,breaklinks,colorlinks]{hyperref}

\usepackage[capitalize]{cleveref}
\crefname{section}{Sec.}{Secs.}
\Crefname{section}{Section}{Sections}
\Crefname{table}{Table}{Tables}
\crefname{table}{Tab.}{Tabs.}

\newenvironment{packed_itemize}{
	\vspace{-0.15cm}\begin{itemize}
		\setlength{\itemsep}{1pt}
		\setlength{\parskip}{0pt}
		\setlength{\parsep}{0pt}
	}{\end{itemize}}

\def\cvprPaperID{3326} 
\def\confName{CVPR}
\def\confYear{2023}

\begin{document}

\title{Generating Aligned Pseudo-Supervision from Non-Aligned Data for \\ Image Restoration in Under-Display Camera}


\author{
Ruicheng Feng$^{1}$\quad
Chongyi Li$^{1}$\quad
Huaijin Chen$^{2}$\quad
Shuai Li$^{2}$\quad
Jinwei Gu$^{3,4}$\quad
Chen Change Loy$^{1}$
\\
$^{1}$S-Lab, Nanyang Technological University\quad
$^{2}$SenseBrain Technology\\
$^{3}$The Chinese University of Hong Kong\quad
$^{4}$Shanghai AI Laboratory\\
{\tt\small \{ruicheng002, chongyi.li, ccloy\}@ntu.edu.sg}
\\
{\tt\small \{huaijin.chen, shuailizju\}@gmail.com\quad
jwgu@cuhk.edu.hk}
}

\maketitle


\begin{abstract}
%
Due to the difficulty in collecting large-scale and perfectly aligned paired training data for Under-Display Camera (UDC) image restoration, previous methods resort to monitor-based image systems or simulation-based methods, sacrificing the realness of the data and introducing domain gaps.
In this work, we revisit the classic stereo setup for training data collection --  capturing two images of the same scene with one UDC and one standard camera. The key idea is to ``copy'' details from a high-quality reference image and ``paste'' them on the UDC image. While being able to generate real training pairs, this setting is susceptible to spatial misalignment due to perspective and depth of field changes. The problem is further compounded by the large domain discrepancy between the UDC and normal images, which is unique to UDC restoration.
In this paper, we mitigate the non-trivial domain discrepancy and spatial misalignment through a novel Transformer-based framework that generates well-aligned yet high-quality target data for the corresponding UDC input. 
This is made possible through two carefully designed components, namely, the Domain Alignment Module (DAM) and Geometric Alignment Module (GAM), which encourage robust and accurate discovery of correspondence between the UDC and normal views. 
Extensive experiments show that high-quality and well-aligned pseudo UDC training pairs are beneficial for training a robust restoration network.
Code and the dataset are available at \url{https://github.com/jnjaby/AlignFormer}.
\end{abstract}

 \vspace{-0.3cm}
\section{Introduction}

\begin{figure}[t]
    \centering
    \includegraphics[width=\linewidth]{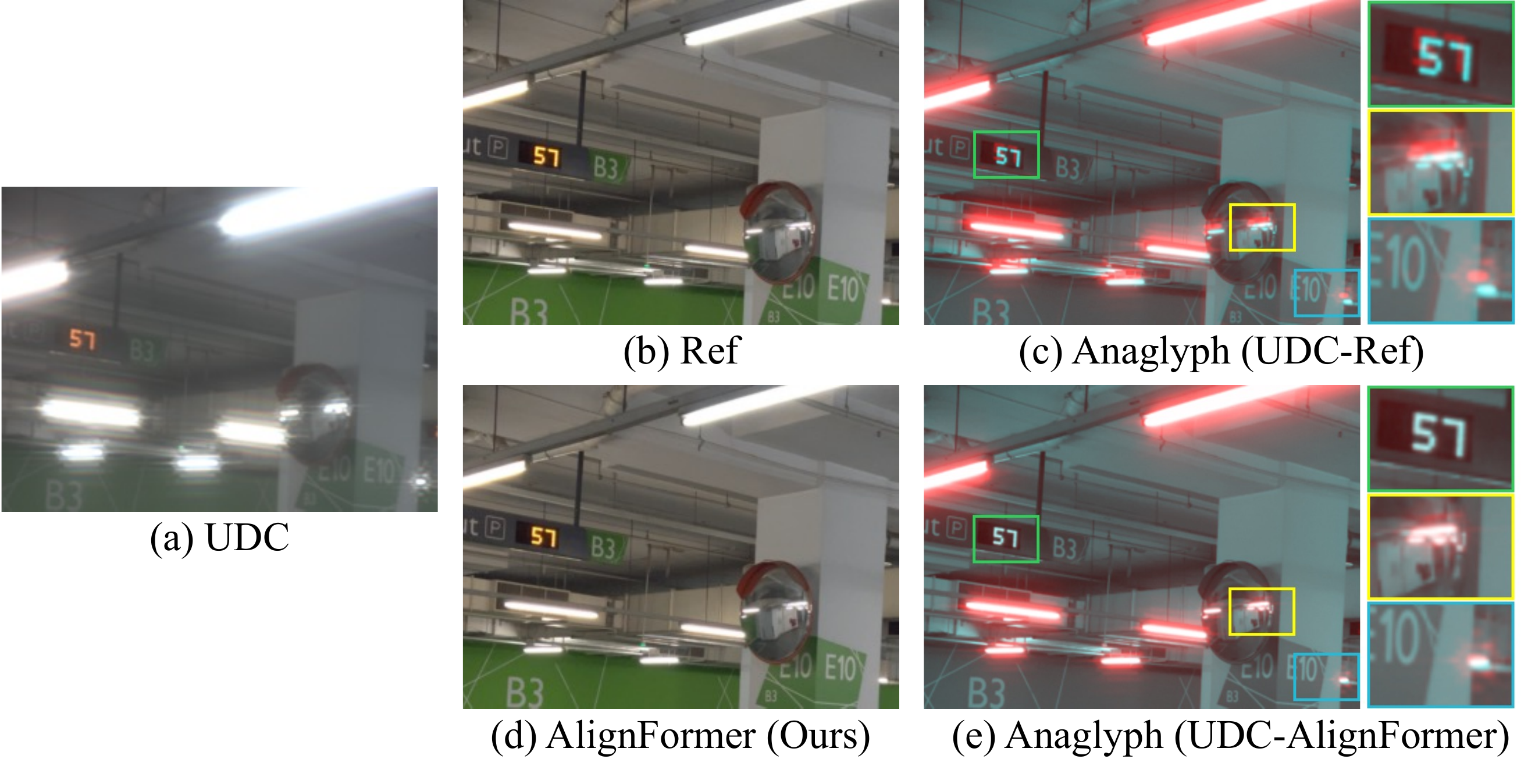}
    \vspace{-0.8cm}
    \caption{\textbf{Domain and geometric misalignment in UDC.}
    Stereo pairs (a) and (b) are captured by Under-Display Camera and high-end camera, respectively. The two images deviate significantly due to the color shift and severe degradation the UDC image. Anaglyph (c) illustrates the large spatial displacement between UDC and reference images despite a careful hardware setup and rough alignment. Our AlignFormer aligns the image pair and minimizes the parallax.
    }
    \label{fig:teaser}
    \vspace{-0.3cm}
\end{figure}

Under-Display Camera (UDC) is an imaging system with cameras placed underneath a display. It emerges as a promising solution for smartphone manufacturers to completely hide the selfie camera, providing a notch-free viewing experience on smartphones.
However, the widespread commercial production of UDC is prevented by poor imaging quality caused by diffraction artifacts.
Such artifacts are unique to UDC, caused by the gaps between display pixels that act as an aperture.
As shown in Figure~\ref{fig:teaser}(a), typical diffraction artifacts entail flare, saturated blobs, blur, haze, and noise. The complex and diverse distortions make the reconstruction problem extremely challenging.

Training a deep network end-to-end for UDC restoration has been found challenging due to the need for a large-scale dataset of real-world degraded images and their high-quality counterparts.
Existing methods~\cite{zhou2021image,qi2021isp} build datasets with a monitor-based imaging system. As discussed in Feng~\etal~\cite{feng2021removing}, such a paradigm is inherently limited by the dynamic range and spatial resolution of the monitor. To address the problem, Feng~\etal~\cite{feng2021removing} present a synthetic dataset grounded on the imaging formation model~\cite{feng2021removing}.
Both datasets exhibit degradation that deviates from the actual physical imaging process, leading to poor generalizability to diverse real-world test cases.

To circumvent the hurdle in collecting real paired data, we opt for an alternative setup, \ie, to construct paired dataset with a \textit{stereo} setting.
Specifically, we capture two images of the same scene with one Under-Display Camera and one normal camera, denoted as UDC and Reference image, respectively. An example is shown in Figure~\ref{fig:teaser}(a-b)
The key challenge lies in two aspects.
i) \textbf{Domain discrepancy.} The different camera configurations inevitably give rise to variations in illuminance and severe color inconsistency, especially under the presence of color shift and severe diffraction artifacts in the UDC image.
ii) \textbf{Geometric misalignment.} The contents in the UDC image and reference image are misaligned due to different focal lengths and field of views (FOV).

Due to the unique nature of UDC restoration, existing solutions are not effective in addressing the two aforementioned challenges. In particular,  the low-level vision community has made attempts on this stereo setup for super-resolution~\cite{cai2019toward}, deblurring~\cite{rim2020real}, and learnable ISP~\cite{ignatov2020replacing}. In addition, Contextual loss \cite{mechrez2018contextual} and CoBi loss \cite{zhang2019zoom} are devised to alleviate mild spatial misalignment. As shown in our experiments, those methods are less stable and robust due to the difficulty of reliable matching when one image is severely distorted. In particular, the over-exposed regions caused by diffraction require strong pixel-wise supervision to enforce constraints during the training. 

The key idea of our solution is to generate high-quality and well-aligned pseudo pairs from the non-aligned stereo data (UDC and reference) to enable end-to-end training of a deep network. The challenge lies in solving the domain and spatial misalignment so that the process resembles `copying' details from the reference image selectively and then `pasting' on the degraded image.
To this end, we devise a simple yet effective Transformer-based framework, namely \textit{AlignFormer}, with a Domain Alignment Module (DAM) and a Geometric Alignment Module (GAM).
The DAM is inspired by AdaIN \cite{huang2017arbitrary}, aiming to mitigate the domain discrepancy between the UDC and reference images, allowing more robust and accurate correspondence matching in the subsequent stage.
The GAM establishes accurate dense correspondences through incorporating geometric cues in attention. 
Specifically, GAM can flexibly work with any off-the-shelf pre-trained optical flow estimators to build pixel-wise correspondence between the UDC and reference images. The discovered correspondence then guides the sparse attention in our Transformer to search for the matching pixels accurately and effectively within local regions.

Figure~\ref{fig:teaser}(d-e) show that AlignFormer produces well-aligned image pairs.
The results of AlignFormer can serve as pseudo ground-truth data and one can easily train an image restoration network end-to-end with common training settings, \ie, using pixel losses such as $\mathcal{L}_1$ that assume exact spatial alignment, the perceptual loss \cite{johnson2016perceptual}, and the adversarial loss.
Moreover, the constructed pseudo-paired dataset allows us to enjoy the merits of any advanced architectures of neural networks designed for image restoration problems.
The generated data do not suffer from the limited dynamic range of spatial resolution as in previous monitor-based imaging systems. The data also experience a far lower domain gap than simulation-based approaches.

The main contributions are three-fold:
\begin{packed_itemize}
    \item We propose a data generation framework that is specifically designed for UDC. It presents a promising direction beyond previous monitor-based and simulation-based data collection approaches, leading to improved generalizability of UDC image restoration.
    \item Our AlignFormer properly integrates optical flow guidance into up-to-date Transformer architectures.
    \item Experimental results demonstrate significant progress in practical UDC image restoration in real-world scenarios.
\end{packed_itemize}


\vspace{-0.3cm}
\section{Related Work}
\noindent{\bf UDC Image Restoration.}
Very few works in the literature have investigated image restoration for UDC.
Zhou \etal~\cite{zhou2021image} and ECCV 2020 challenge \cite{zhou2020udc} pioneered this line of works and inspired the follow-up studies \cite{feng2021removing,kim2021under,kwon2021controllable,oh2021residual}.
Yang \etal~\cite{yang2021designing} proposed to redesign the pixel layouts for UDC display by optimizing the display patterns to improve the quality of restored image, which is orthogonal to our work.
The dataset of the challenge \cite{zhou2020udc,zhou2021image} is captured by a monitor-based imaging system. Such a system only induces incomplete diffraction artifacts due to the limited dynamic range of monitor \cite{feng2021removing}.
Qi \etal~\cite{qi2021isp} further explored the use of HDR monitor data. However, it is still inherently limited by the spatial resolution and contrast of the monitor.
To remedy this issue, instead of capturing monitor-based image pairs, Feng \etal~\cite{feng2021removing, feng2022mipi} and Gao \etal~\cite{gao2021image} explored simulation pipelines for building synthetic dataset with real-captured point spread function (PSF) and imaging formation model.
Despite well calibration and correction of PSF, models trained on synthetic dataset exhibit limited generalization capability for real-world images, especially for those with strong illuminations and flare regions.
This is partially due to the domain shift between the mathematical model and physical imaging process in the real world, as there is no guarantee that the simulation pipeline can well approximate the complicated degradation in practical scenarios.
Unlike previous works, we propose to collect real-world degraded-reference image pairs and produce high-quality images that contain the same content as degraded images as the pseudo target.
The pseudo label allows us to enjoy the merits of advanced network architectures that are trained with pixel-wise losses.

\noindent{\bf Dealing with Misaligned Paired Data.}
Several studies have been devoted to capturing real-world image pairs using different cameras or camera configurations for other low-level tasks.
Qu \etal~\cite{qu2016capturing} and Rim \etal~\cite{rim2020real} devised image acquisition systems with a beam splitter to collect paired data.
Wang \etal~\cite{wang2021DCSR} present dual-camera super resolution.
Ignatov \etal~\cite{ignatov2020replacing} and Zhang \etal~\cite{zhang2021learning} collected image pairs and roughly aligned them via SIFT keypoints \cite{lowe2004distinctive} and RANSAC algorithm \cite{vedaldi2010vlfeat}. The geometric alignment algorithm they adopted assumes image pairs can be aligned with a single homography, which is not the case where depth discrepancy exists in the scenes.
Cai \etal~\cite{cai2019toward} developed a pixel-wise image registration method to iteratively transform and adjust luminance. Nonetheless, this algorithm only works while the misalignment is mild.
Contextual loss (CX) \cite{mechrez2018contextual} was proposed to relax the constraints of spatial alignment and the loss works based on context and semantics.
Inspired by CX \cite{mechrez2018contextual}, Zhang \etal~\cite{zhang2019zoom} presented a contextual bilateral (CoBi) loss to prioritize local features and improve the matching quality.
None of them consider image pairs exhibiting severe degradation and color inconsistency.
%
%
Another line of studies focuses on transferring textures and details from the reference image, \eg, TTSR~\cite{yang2020learning}, $\mathcal{C}^2$-matching~\cite{jiang2021robust,jiang2022reference}. Our framework is inspired by these prior studies. These algorithms only discover pixel correspondence through semantic similarity without using any geometric cues. Our approach differs in guiding the Transformer's attention with geometric prior and considering the additional domain alignment for addressing the significant gaps between the UDC and reference images.

\section{Method}
Given a UDC image $I_D\in\mathbb{R}^{H\times W\times 3}$ and a high-quality reference image $I_R\in\mathbb{R}^{H\times W\times 3}$ of the same scene, we aim to generate $I_P$ that possesses finer texture and details from $I_R$ and preserves the content of $I_D$,
\begin{equation}
    I_P=\mathcal{T}(I_D,I_R;\Theta),
\end{equation}
where $\mathcal{T}$ denotes the transformation.
The whole process can be regarded intuitively as ``copying'' texture from reference images and ``pasting'' to the target image according to the semantic content of degraded images.
The constructed pseudo pair $(I_D,I_P)$ is well-aligned and could serve as a training sample to provide better supervision for subsequent UDC image restoration networks $f_{\theta}$, given by
\begin{equation}
    I_O = f_{\theta}(I_D),
\end{equation}
where $I_O$ is the reconstructed clean image.
The key challenge lies in aligning $I_R$ to $I_D$ in the presence of significant domain inconsistency.
To overcome this, we propose a novel Transformer-based framework, \textit{AlignFormer}, to mitigate both domain shift and spatial misalignment.

\begin{figure}[t]
    \centering
    \includegraphics[width=.9\linewidth]{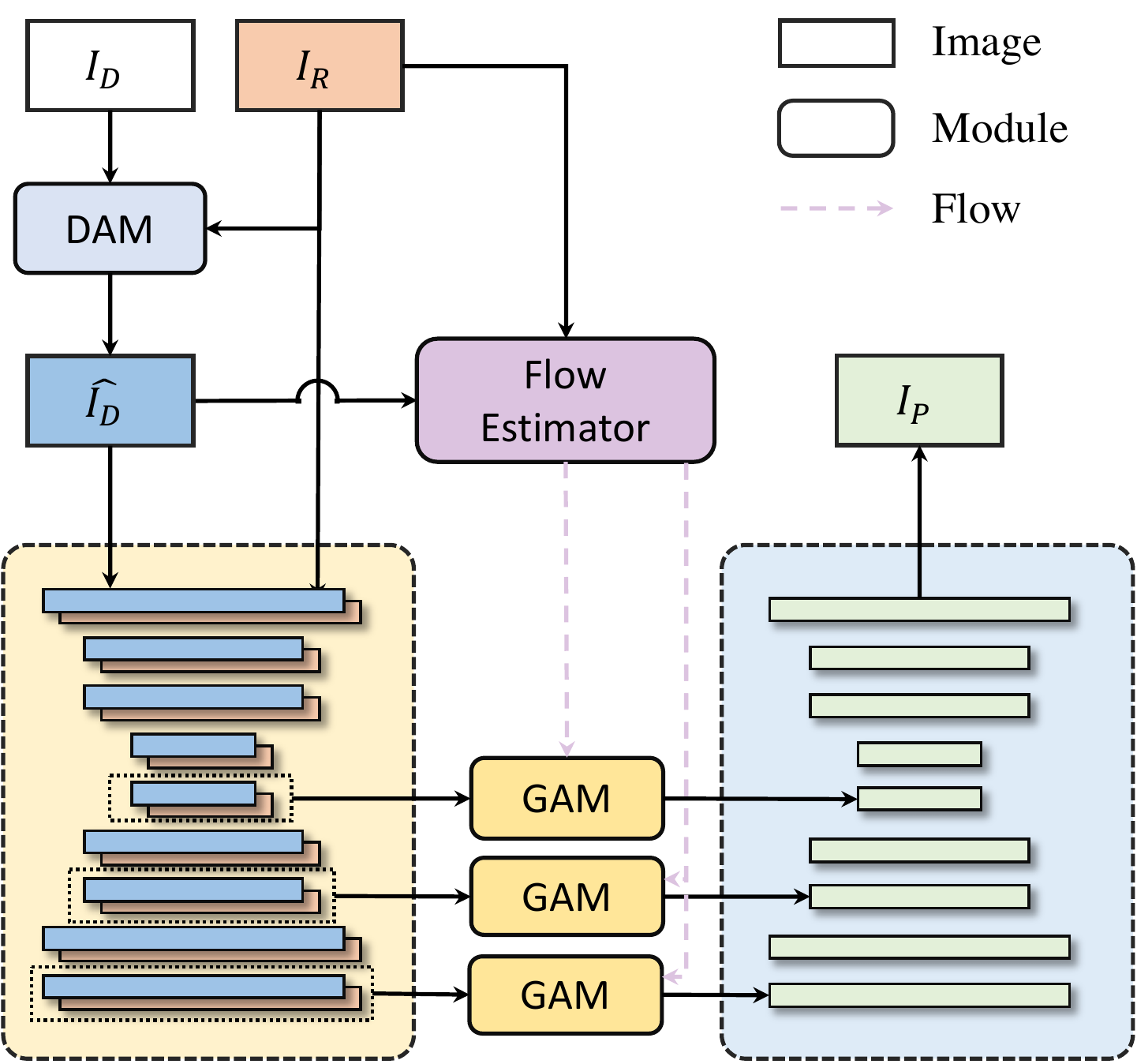}
    \vspace{-0.2cm}
    \caption{\textbf{Overview of the proposed AlignFormer.}
    We first mitigate domain discrepancy between UDC image $I_D$ and reference image $I_R$ via Domain Alignmain Module (DAM) to obtain $\hat{I_D}$, which are then gathered with $I_R$ and fed into two U-shape CNNs for feature extraction. Then the features at each scale are attended by the Geometric Alignment Transformer (GAM) to obtain the output features, which will be processed and fused in another U-Net to produce the pseudo image $I_P$.
    }
    \label{fig:architecture}
    \vspace{-.5cm}
\end{figure}

\subsection{AlignFormer}
Inspired by Texformer~\cite{xu20213d}, the overall architecture of AlignFormer $\mathcal{T}$ is illustrated in Fig.~\ref{fig:architecture}.
It mainly consists of Domain Alignment Module (DAM), Geometric Alignment Module (GAM), flow estimator, and feature extractors.
The DAM is carefully designed to modulate features towards reduced domain inconsistency, and consequently improve the accuracy of correspondence matching.
On top of it, the GAM establishes accurate dense correspondences by incorporating geometric cues derived from any off-the-shelf optical flow estimators in attention. This enables sparse attention in our Transformer to search for the matching pixels accurately and effectively within local regions.

\noindent\textbf{Domain Alignment Module (DAM).}
\begin{figure}[t]
    \centering
    \includegraphics[width=.85\linewidth]{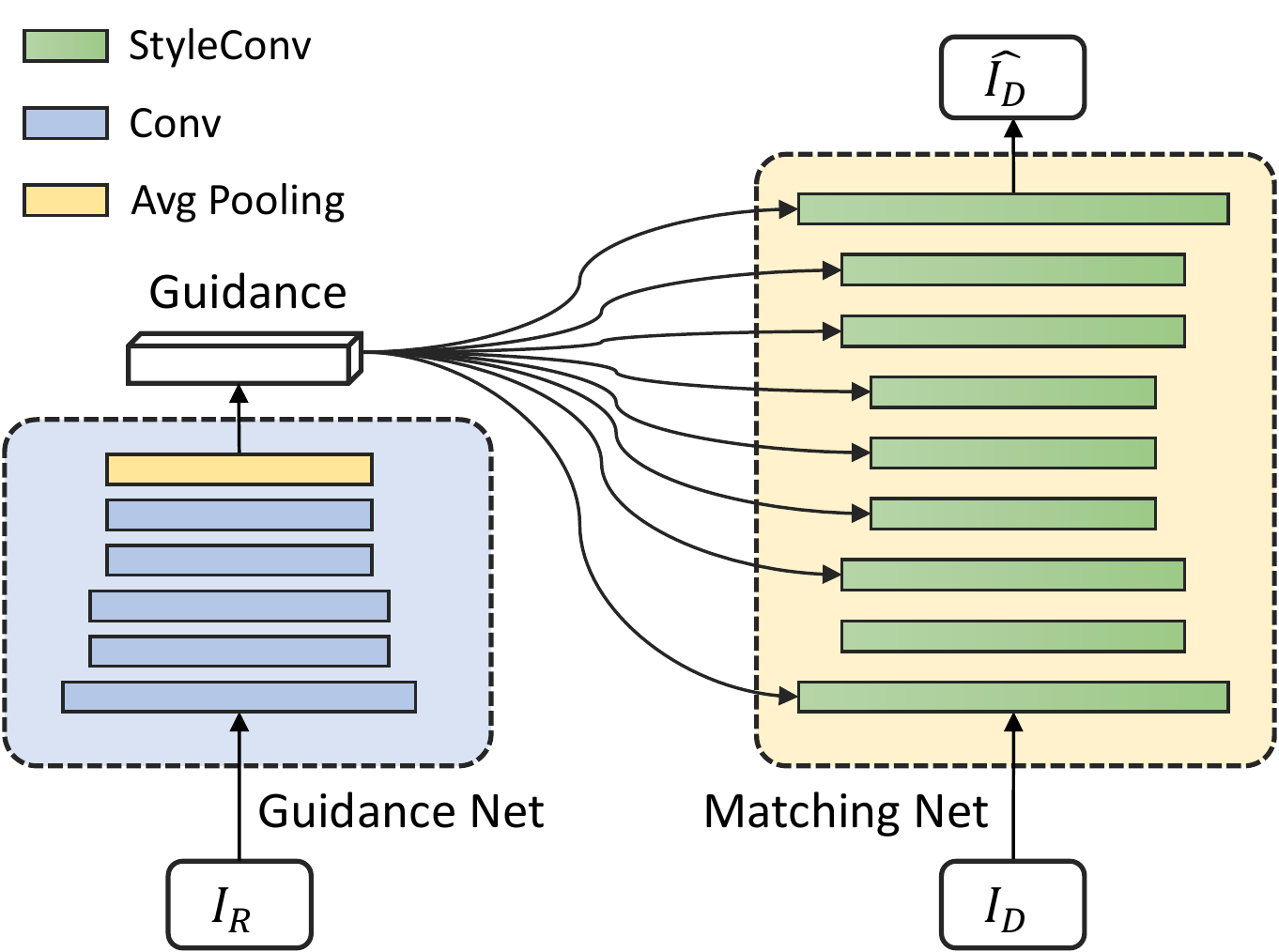}
     \vspace{-0.2cm}
    \caption{\textbf{The structure of domain alignment module.}
    This module comprises a guidance net and a matching net. The guidance vector generated by the guidance net is used for style modulation via StyleConv in the matching net. Such designs help to imitate the color and illuminance of reference image, while preserving spatial information of UDC image.
    }
    \label{fig:dam}
     \vspace{-0.4cm}
\end{figure}
From our experiments, we found that the performance of attention is highly susceptible to domain shift and severe degradations.
Inspired by the recent success of style-based architecture~\cite{karras2019style,karras2020analyzing}, we propose DAM to mitigate domain inconsistency between the UDC and reference images.
The structure of DAM is shown in Fig.~\ref{fig:dam}, consisting of two sub-nets, a guidance net and a matching net. 

The guidance net generates a conditional vector as guidance for the matching net by extracting and exploring feature statistics, \ie, domain information, from the reference image $I_R$. To obtain the guidance vector, denoted as $\vb*{s}\in\mathbb{R}^d$, we compose a stack of convolution layers, followed by a global average pooling layer.
The guidance vector can serve as the condition and deliver holistic information from the reference image to the matching net.

The matching net is designed to transfer domain information, \eg, color, illuminance, contrast, to the degraded UDC image.
To leverage the style condition $\vb*{s}\in \mathbb{R}^d$ and match the feature, we utilize StyleConv~\cite{karras2019style}, consisting of a Conv layer, affine transformation, and AdaIN~\cite{huang2017arbitrary}. Specifically, let $\mathbf{A}\in \mathbb{R}^{2d\times d}$ and $\mathbf{b}\in \mathbb{R}^{2d}$ be the layer-wise learnable affine transformation applied to $\vb*{s}$.
For each AdaIN layer, we can define the style input as $\vb*{y}=[\vb*{y}_s, \vb*{y}_b]=\mathbf{A}\vb*{s}+\vb*{b}$.
Given the output feature of Conv layer $\vb*{x}$ as the content input and style input $\vb*{y}$, AdaIN performs 
\begin{equation}
    \text{AdaIN}(\vb*{x},\vb*{y})=
    \vb*{y}_s \frac{\vb*{x}-\mu(\vb*{x})}{\sigma(\vb*{x})} + \vb*{y}_b.
\end{equation}
The output of DAM, denoted as $\hat{I_D}$, will exhibit a closer style to the $I_R$, and will be used for the subsequent GAM. Note that the style transfer is performed from $I_R$ to $I_D$, and not otherwise, as the latter will hamper the correspondence discovery due to poorer quality of image pairs.

\noindent\textbf{Geometric Alignment Module (GAM).}
\begin{figure}[t]
    \centering
    \includegraphics[width=.9\linewidth]{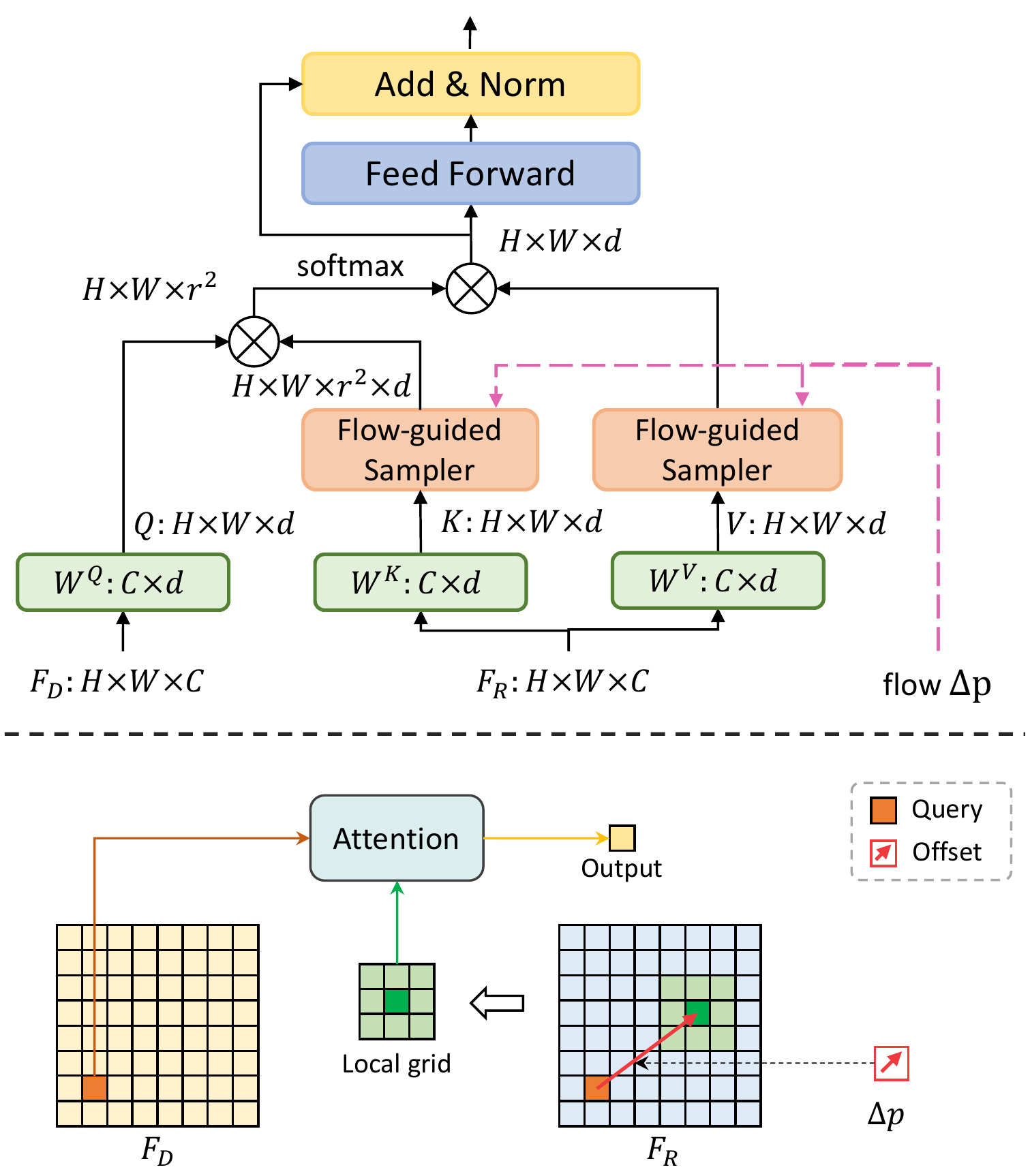}
    \vspace{-0.3cm}
    \caption{Top: The structure of geometric alignment module.
    Bottom: the schematic of flow-guided sampler.}
    \vspace{-0.4cm}
    \label{fig:gam}
\end{figure}
The conventional formulation in Transformers~\cite{vaswani2017attention,dosovitskiy2021vit,liu2021Swin} is not well-suited for our task.
In particular, the vanilla Transformers~\cite{vaswani2017attention,dosovitskiy2021vit} exhibit the superiority in exploiting semantic similarity and capturing global contextual information. Nonetheless, their global attention densely attends to all key components, causing a diversion of attention to irrelevant and redundant elements.
Swin Transformer~\cite{liu2021Swin}, on the other hand, assumes attention within local regions, which is better suited for our task. However, it is not readily designed to accept geometric cues (\eg, optical flow, patch matching) as guidance, which is crucial in our task considering local rigidity assumption under the stereo setting.
Hence, to discover correspondences between the UDC input $I_D$ and reference image $I_R$, a specialized attention mechanism is required.

To this end, we develop GAM to exploit the pixel correspondences by incorporating geometric cues in attention.
Specifically, we use an off-the-shelf flow estimator\footnote{We use RAFT~\cite{teed2020raft} but other methods are applicable.} as the guidance to sample features from the vicinity in the reference image.
We introduce a flow estimator prior in the conventional attention as it can exploit the geometric prior to facilitate the subsequent attention mechanism. Otherwise, it becomes knotty given the spatial misalignment between $\hat{I_D}$ and $I_R$.
As shown in Figure~\ref{fig:gam}, we reformulate the basic attention unit.
It consists of attention with a flow-guided sampler,
a layer normalization (LN),
and a multi-layer perceptron (MLP).

%
To take advantage of the geometric cues,
the \textit{key} sampling is directed by the geometric information predicted by the pre-trained optical flow estimator.
The dense offset map at each position $\mathbf{p}=(x,y)$ in $\hat{I_D}$ is mapped to its estimated correspondence in $I_R$: $\mathbf{p}'=(x+u, y+v)$, where the flow offset is recovered by a network $\psi_{flow}$
\begin{equation}
    \Delta \mathbf{p} = (u,v) = \psi_{flow}(\hat{I_D},I_R)(x,y).
\end{equation}
Then a flow sampler defines the local grid around $\mathbf{p}'$
\begin{equation}
    \mathcal{N}(\mathbf{p}')_r = \{\mathbf{p}'+\mathbf{d}\mid\mathbf{d}\in\mathbb{Z}^2,\|\mathbf{d}\|_1\leq r\}.
\end{equation}
The offsets within a radius of $r$ units are used to sample key and value elements.
The flow sampler samples sub-pixel locations of real values from $\mathcal{N}(\mathbf{p}')_r$ via  interpolation. 

We denote $F_D\in \mathbb{R}^{H\times W\times C}$ and $F_R\in \mathbb{R}^{H\times W\times C}$ the feature extracted from $\hat{I_D}$, $I_R$, where $H$ and $W$ indicate height and width of the feature map, and $C$ is the feature dimension.
Given a linear projected query vector $\vb*{q}_{x,y}=F_D W^Q$ at coordinate $\mathbf{p}=(x,y)$ of $F_D$, the flow-guided attention can be written as:
\begin{equation}
    f_{attn}(\vb*{q}_{x,y}, F_R) = 
    \sum_{(i,j)\in \mathcal{N}(\mathbf{p}')_r}
    \text{sim}(\vb*{q}_{x,y},\vb*{k}_{i,j}) \vb*{v}_{i,j},
\end{equation}
where $\vb*{k}_{i,j}=F_R W^K$ and $\vb*{v}_{i,j}=F_R W^V$ represent the projected vectors sampled from $F_R$ by the flow sampler.
Here $W^Q$, $W^K$, $W^V\in \mathbb{R}^{C\times d}$ are the respective learnable linear projection for query, key, and value elements,
where $d$ is the dimension of the projected vector.
The attention score $\text{sim}(\vb*{q}_{x,y},\vb*{k}_{i,j})$ is the scaled dot-product attention followed by softmax function, formulated as

\begin{equation}
    \text{sim}(\vb*{q}_{x,y},\vb*{k}_{i,j}) = \text{softmax} (\frac{\vb*{q}_{x,y}^\mathsf{T} \vb*{k}_{i,j}}{\sqrt{d}}).
\end{equation}
Hence, the final output attended features are computed as

\begin{equation}
    \begin{array}{l}
        \vb*{f}_{x,y} = f_{attn}(\vb*{q}_{x,y}),\\
        \vb*{z}_{x,y} = f_{\text{MLP}}(f_{\text{LN}}(\vb*{f}_{x,y}))+\vb*{f}_{x,y}.
    \end{array}
\end{equation}
Here $f_{\text{LN}}$ is the LayerNorm layer.

\subsection{Learning Objectives}
The training of AlignFormer requires an objective function that does not forcefully match each spatial position as the problem lacks exact spatially aligned supervision.
The Contextual Loss (CX) \cite{mechrez2018contextual} is a viable choice as it treats features of images as a set and measures the similarity between images, ignoring the spatial positions of the features.
This property enables us to compare images that are spatially deformed. 

Given two images $x$ and $y$, CX loss aims to minimize the summed distance of all matched feature pairs, formulated as
\begin{equation}
    \mathcal{L}_{CX}(x,y) = \frac{1}{N}
    \sum_j \min_i
    \mathbb{D}(\phi(x)_j, \phi(y)_i),
\end{equation}
where $\phi(x)_j$ and $\phi(y)_i$ are the $j$-th point of $\phi(x)$ and $i$-th point of $\phi(y)$, respectively. $\phi(x)$ denotes feature maps of $x$ extracted from the VGG network $\phi$, and $\mathbb{D}$ is some distance measure. 
Based on context and semantics, the CX loss transfers the style of an image to another by comparing regions with similar semantic meaning in both images. The insensitivity of CX loss for misaligned data is well-suitable for moderating the domain gap between UDC images and high-quality reference images used in our pipeline.

\noindent\textbf{Learning Objective for DAM.}
We first train DAM to mitigate the domain shift. The loss term for DAM is given by
\begin{equation}
    \mathcal{L}_{DAM}=\mathcal{L}_{CX}(\hat{I_D},I_R),
\end{equation}
where $\hat{I_D}$ denotes the output of DAM, $I_D$ and $I_R$ the degraded and reference image, respectively.
We use the pretrained VGG-19 \cite{simonyan2014very} and select ``conv4\_4'' as deep features.

\noindent\textbf{Learning Objective for AlignFormer.}
After training DAM, we integrate the DAM and the pre-trained RAFT \cite{teed2020raft} into the AlignFormer. Note that both DAM and the optical flow estimator are fixed during training AlignFormer.
Similarly, the AlignFormer is trained with a domain loss:
\begin{equation}
    \mathcal{L}_{Align}=\mathcal{L}_{CX}(I_P,I_R),
\end{equation}
where $I_P$ is the output of AlignFormer. The reference image $I_R$ serves as the domain supervision.


\subsection{Image Restoration Network}
Although AlignFormer can achieve good results with a UDC image and a reference image as input, a restoration network is still indispensable.
This is because the reference image is not available during inference and some regions in AlignFormer's results have to be masked out due to the effect of occlusion between the UDC image and the reference image. 
Therefore, after getting the aligned pseudo image pairs ($I_D$, $I_P$), we devise a baseline network targeting at restoring both global information (\eg, brightness and color correction), and local information (\eg, flare removal, texture enhancement, denoising) to evaluate the effectiveness of our AlignFormer.
While a tailored image restoration backbone UNet~\cite{ronneberger2015u} could inherently enhance local details at different scales, it can hardly alter the image globally.
To address this, we adopt a standard U-Net architecture with Pyramid Pooling Module (PPM)~\cite{zhao2017pyramid}, namely \textit{PPM-UNet}.
The role of PPM is to incorporate global prior into the networks to mitigate the color inconsistency between UDC and generated images. 
Due to limited space, we leave the details of the restoration network to the supplementary material.

\noindent\textbf{Learning Objective for PPM-UNet.}
Following common practice \cite{zhang2021learning,wang2018esrgan}, we train PPM-UNet with a combination of losses, including $\mathcal{L}_1$ loss, VGG-based perceptual loss \cite{johnson2016perceptual} $\mathcal{L}_{VGG}$, and GAN loss, which can be defined by
\begin{equation}
\begin{split}
    \mathcal{L}_{rec}=&\lambda_1 \|M\odot (I_P-I_O)\|_1\\
    +&\lambda_{VGG} \|\phi(M\odot I_P) - \phi(M\odot I_O)\|_1\\
    +&\lambda_{GAN} \mathcal{L}_{GAN},
\end{split}
\end{equation}
where $\odot$ denotes element-wise multiplication, $\|\cdot\|_1$ is $\ell_1$-norm, $\phi$ is the pre-trained VGG \cite{simonyan2014very} network, and $M$ is the valid mask indicating the non-occluded regions of optical flow.
To avoid deteriorating networks due to inaccurate deformations over occluded regions, we mask out those pixels invisible in the reference image.
The occlusion detection is derived from forward-backward consistency assumption~\cite{sundaram2010dense}.
To further improve the visual quality, we also add adversarial loss based on conditional PatchGAN~\cite{isola2017image}.
Please refer to the supplementary material for details.


\begin{table*}[t]
\caption{\textbf{Comparison between different datasets on the baseline network.}
	We train the PPM-UNet using different training sets and evaluate their performance on the test set.
	Note that for non-aligned data, we also use CX~\cite{mechrez2018contextual} loss and CoBi~\cite{zhang2019zoom} loss to tackle with misalignment.
	The best and runner up results are highlighted in \textbf{bold} and \underline{underlined}, respectively.
}
\label{tab:dataset_comparison}
\vspace{-0.3cm}
\centering
\resizebox{.8\textwidth}{!}{
\begin{tabular}{ccc|ccc|cc|ccc}
\toprule[0.1em]
\multirow{2}{*}{Training sets} & \multirow{2}{*}{Aligned} & \multirow{2}{*}{Loss} & \multicolumn{3}{c|}{\textit{Aligned Ref}} & \multicolumn{2}{c|}{\textit{Original Ref}} & \multicolumn{3}{c}{\textit{Non Ref}}\\
 & & & PSNR$\uparrow$ & SSIM$\uparrow$ & LPIPS$\downarrow$ & CD (L / a / b)$\uparrow$ & SIFID {\tiny ($\times 10^{-5}$)}$\downarrow$ & NIQE$\downarrow$ & MUSIQ$\uparrow$ & NRQM$\uparrow$ \\
\midrule
Synthetic~\cite{feng2021removing} & \checkmark & $\mathcal{L}_{rec}$ & 19.03 & 0.7808 & 0.3513 & 0.67 / 0.40 / 0.27 & 6.3341 & \underline{6.4706} & 34.0738 & 4.8640 \\
\midrule
Real & & $\mathcal{L}_{rec}$ & 19.04 & 0.8187 & 0.1979 & 0.93 / \underline{0.46} / \underline{0.47} & 2.2157 & 7.5641 & \underline{53.1251} & 5.8763 \\
Real & & $\mathcal{L}_{CX}$ & 20.85 & 0.8198 & 0.1524 & 0.93 / 0.25 / 0.40 & 1.1508 & 9.7242 & 48.8314 & 5.9143 \\
Real & & $\mathcal{L}_{CoBi}$ & \underline{21.57} & \underline{0.8319} & \underline{0.1252} & \underline{0.93} / 0.30 / 0.41 & \underline{1.0385} & 8.9563 & 50.8363 & \underline{6.0025} \\
\midrule
Real & AlignFormer & $\mathcal{L}_{rec}$ & \textbf{22.95} & \textbf{0.8581} & \textbf{0.1236} & \textbf{0.94} / \textbf{0.48} / \textbf{0.47} & \textbf{0.9735} & \textbf{6.2816} & \textbf{56.3314} & \textbf{6.4839}\\
\bottomrule[0.1em]
\end{tabular}
}
\vspace{-0.3cm}
\end{table*}
\begin{figure*}[t]
    \centering
    \includegraphics[width=.9\linewidth]{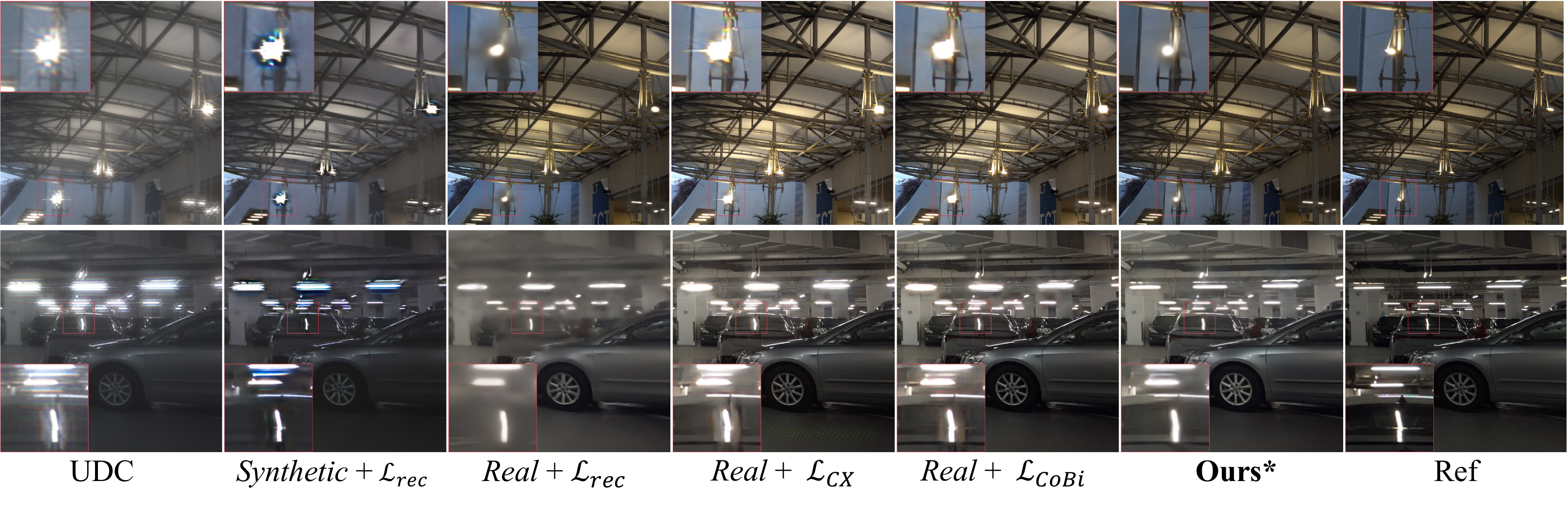}
    \vspace{-0.4cm}
    \caption{\textbf{Visual comparison between different datasets on the baseline network.}
    * indicates results of ``AlignFormer + PPM-UNet''}
    \label{fig:dataset_comparison}
    \vspace{-0.5cm}
\end{figure*}


\section{Experiments}
\noindent\textbf{Data Collection and Pre-Processing.}
\label{sec:dataset}
To build the degraded-reference image pairs, we construct a stereo smartphone array - ZTE Axon 20 with selfie under-display camera, and iPhone 13 Pro rear camera, which are physically aligned as much as possible (see supplementary material for details).
To eliminate the effects of built-in ISPs, both UDC and Ref images are extracted from raw dump of data with minimal processing (demosaic and gamma correction) and converted into sRGB domain.
In total, we collect $330$ image pairs covering both indoor and outdoor scenes.

Given a pair of images, we first roughly align them using a homography transformation estimated via RANSAC algorithm \cite{vedaldi2010vlfeat} as it is robust to photometric misalignment.
Even after alignment with homography, there still exists mild misalignment between the image pair. 
This is because the geometric transformation is applied globally with the assumption that all points are located at the same plane in the world, which does not hold true where contents are at different depths in the scene.
The remaining displacement in the pair will be further resolved by our AlignFormer.

\noindent\textbf{Implementation.}
We split $330$ image pairs into $274$ pairs for training and $56$ for testing.
For each pair of full-resolution images, we crop them into $512\times512$ patches for training. More example training patches can be found in the supplement.
We initialize all networks with Kaiming Normal \cite{he2015delving} and train them using Adam optimizer \cite{kingma2014adam} with $\beta_1=0.9$, $\beta_2=0.999$ and $\theta=10^{-8}$, and the mini-batch size is set to $8$ for all the experiments.
$\lambda_1$, $\lambda_{VGG}$, and $\lambda_{GAN}$ are set to $10^{-2}$, $1$, $5\times 10^{-3}$, respectively.
The learning rate is decayed by half at $250k$ and $300k$ iterations with a multi-step schedule. We implement our models with PyTorch \cite{paszke2017automatic} and train them using two NVIDIA V100 GPU cards.

\subsection{Comparisons}
\paragraph{Evaluation Protocol.}
In the absence of ground truths with spot-on alignment, it is challenging to  evaluate the quality of generated images.
To quantify the performance, we compute three common metrics (\eg, PSNR, SSIM, and LPIPS~\cite{zhang2018unreasonable}) on the Pseudo GT $I_P$, namely \textit{Aligned Ref}.
We additionally adopt two sets of evaluations for a comprehensive and accurate comparison.
To quantify the realism of the generated images and how well they capture the domain information (\eg, color, illuminance, contrast) of the reference images (\textit{Original Ref}),
we use two measures: CD - Color Distribution~\cite{isola2017image}, and SIFID~\cite{shaham2019singan} - Single Image FID.
Specifically, color distribution is to calculate the histogram intersection of marginal color distributions between our results and reference images in Lab color space.
SIFID is to measure the internal statistics of patch distributions of single image.
In addition, we adopt two non-reference (\textit{Non Ref}) metrics MUSIQ~\cite{ke2021musiq}, NIQE~\cite{mittal2012making}, and NRQM~\cite{ma2017learning} to supplement the quantitative comparision.

\begin{table*}[t]
\caption{\textbf{Benchmark of state-of-the-art UDC image restoration methods on real dataset.}
	``*'' indicates checkpoint models released by the original paper. Others are retrained with our dataset.
        The best and runner up results are highlighted in \textbf{bold} and \underline{underlined}, respectively.
}
\label{tab:sota_comparison}
\vspace{-0.3cm}
\centering
\resizebox{.8\textwidth}{!}{
\begin{tabular}{l|ccc|cc|ccc}
\toprule[0.1em]
\multirow{2}{*}{\textbf{Method}} & \multicolumn{3}{c|}{\textit{Aligned Ref}} & \multicolumn{2}{c|}{\textit{Original Ref}} & \multicolumn{3}{c}{\textit{Non Ref}}\\
& PSNR$\uparrow$ & SSIM$\uparrow$ & LPIPS$\downarrow$ & CD (L / a / b)$\uparrow$ & SIFID {\tiny ($\times 10^{-5}$)}$\downarrow$ & NIQE$\downarrow$ & MUSIQ$\uparrow$ & NRQM$\uparrow$ \\
\hline
DISCNet*~\cite{feng2021removing} &          19.25 & 0.7574 & 0.3836 & 0.72 / 0.43 / 0.27 & 5.6081 & \underline{6.0847} & 28.0298 & 5.9767 \\
BNUDC*~\cite{koh2022bnudc} &                19.42 & 0.7502 & 0.3574 & 0.73 / 0.40 / 0.28 & 4.9629 & 6.9921 & 30.3077 & \underline{6.5138} \\
\hline
MUNIT~\cite{huang2018munit} &               20.61 & 0.8101 & 0.3258 & \underline{0.93} / 0.44 / 0.41 & 3.1513 & 7.1354 & 36.2816 & 5.5094 \\
TSIT~\cite{jiang2020tsit} &                 19.09 & 0.7755 & 0.2167 & 0.69 / 0.41 / 0.42 & 2.1232 & \textbf{4.6044} & 48.8685 & \textbf{6.6491} \\
\hline
AlignFormer + DISCNet~\cite{feng2021removing} & \underline{21.66} & \textbf{0.8582} & \underline{0.1452} & 0.90 / \underline{0.47} / \underline{0.46} & \underline{1.1623} & 6.7404 & \textbf{56.3539} & 6.0486 \\
AlignFormer + BNUDC~\cite{koh2022bnudc} &   20.43 & 0.8430 & 0.1589 & 0.87 / 0.45 / 0.45 & 2.4669 & 6.5180 & 55.1308 & 5.9677 \\
AlignFormer + PPM-UNet (Ours) &             \textbf{22.95} & \underline{0.8581} & \textbf{0.1236} & \textbf{0.94} / \textbf{0.48} / \textbf{0.47} & \textbf{0.9735} & 6.2816 & \underline{56.3314}& 6.4839\\
\bottomrule[0.1em]
\end{tabular}
}
\vspace{-0.3cm}
\end{table*}
\begin{figure*}[t]
    \centering
    \includegraphics[width=.85\linewidth]{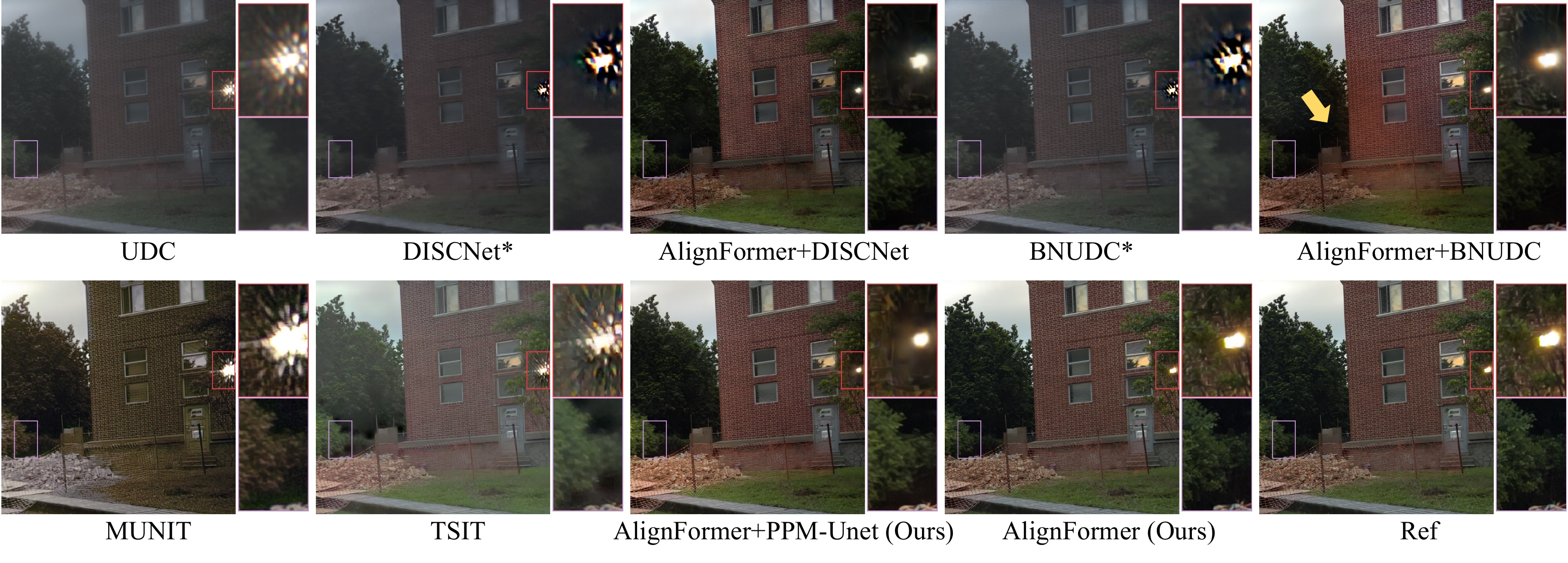}
    \vspace{-0.4cm}
    \caption{\textbf{Visual comparison of benchmarks and our method.}
    ``AlignFormer+'' indicates methods trained on pseudo image pairs generated by AlignFormer.
    Methods marked with * are pre-trained on synthetic dataset.
    }
    \label{fig:sota_comparison}
    \vspace{-0.6cm}
\end{figure*}

\noindent\textbf{Dataset Comparison.}
Before benchmarking existing approaches, we first conduct experiments to compare the performance of our datasets with other synthetic datasets~\cite{feng2021removing}.
Specifically, we train the image restoration network, PPM-UNet, of the same architecture, using several possible combinations of different training sets and learning objectives.
Then, we evaluate their performance on the real test sets and summarize in Table~\ref{tab:dataset_comparison}.
The column ``Aligned'' indicates whether the image pairs are aligned.
The table shows that models trained on the real dataset (even without any alignment) achieve higher performance on our test sets compared to the synthetic dataset in general, which proves that the synthetic dataset is not realistic enough to cover real-world flare images.
Due to the considerable domain gap between synthetic and real data, the model trained on the synthetic dataset (1st row) significantly deteriorates performance on the real data, which validates the necessity of capturing real-world data.
Besides, our baseline network trained on pseudo pairs (last row) demonstrates superior performance against all other methods.
It also achieves the highest PNSR and SSIM scores when compared with the models trained using the misaligned UDC-Reference image pairs regardless of the use of reconstruction loss, CX loss, and CoBi loss.
The quantitative comparisons suggest that the well-aligned and high-quality image pairs are indispensable for the training of UDC image restoration network, which can be achieved by our AlignFormer.
We also show the visual comparisons in Figure~\ref{fig:dataset_comparison}.
We can see that the model trained using the synthetic image pairs still cannot produce a satisfactory result, especially the overall perceptual quality and regions around the saturation.
Moreover, pixel-wise supervision on misaligned image pairs (3rd column) incentivizes blurry results and severe artifacts.
Although CX loss and CoBi loss are devised to alleviate inaccurate alignment, they fail to produce flare-free images and exhibit a different style (\eg, brightness) compared to reference images.
This stems from the absence of strong spatial constraints on flare regions and they are usually performed in feature space.
In contrast, our AlignFormer effectively copies the fine details and textures of the reference image and pastes them back while maintaining alignment with the UDC image.
Such high-quality and well-aligned image pairs, in turn, leading to the visually pleasing result when the model is trained using them, as the ``\textbf{Ours}'' shown in Figure~\ref{fig:dataset_comparison}.

\begin{table*}[ht]
 \centering
 \caption{\textbf{Ablation experiments on AlignFormer.} We report PCK (\%)$\uparrow$ with $\alpha=0.01, 0.03, 0.10$. Our settings are marked in \colorbox{lightgray!50}{gray}.}
\vspace{-0.4cm}
\subfloat[Ablation study on alignment method.]{
    \label{tab:alignment}
    \hfill
    \resizebox{.37\linewidth}{!}{
     \begin{tabular}{c|cccc}
   \toprule[0.1em]
    Alignment & Ref & Cai~\etal~\cite{cai2019toward} & RAFT~\cite{teed2020raft} & \cellcolor{lightgray!50}AlignFormer
    \\\midrule
    $\alpha=0.01$ & 21.77 & 26.15 & 56.14 & \cellcolor{lightgray!50}58.75 \\
    $\alpha=0.03$ & 62.02 & 67.02 & 94.79 & \cellcolor{lightgray!50}95.08 \\
    $\alpha=0.10$ & 85.38 & 79.61 & 98.74 & \cellcolor{lightgray!50}99.93 \\
    \bottomrule[0.1em]
    \end{tabular}}
}
\hfill
\subfloat[Ablation study on optical flow estimator.]{
    \label{tab:exp_optical}
    \hfill
    \resizebox{.37\linewidth}{!}{
     \begin{tabular}{c|cccc}
   \toprule[0.1em]
    Estimator & Zero & SPyNet~\cite{ranjan2017optical} & PWC-Net~\cite{sun2018pwc} & \cellcolor{lightgray!50}RAFT~\cite{teed2020raft}
    \\\midrule
    $\alpha=0.01$ & 9.61 & 31.31 & 33.66 & \cellcolor{lightgray!50}58.75 \\
    $\alpha=0.03$ & 26.86 & 57.45 & 93.22 & \cellcolor{lightgray!50}95.08 \\
    $\alpha=0.10$ & 36.30 & 67.60 & 98.49 & \cellcolor{lightgray!50}99.93 \\
    \bottomrule[0.1em]
     \end{tabular}}
}
\hfill
\subfloat[Ablation study on radius.]{
    \label{tab:exp_radius}
    \hfill
    \resizebox{.21\linewidth}{!}{
     \begin{tabular}{c|ccc}
   \toprule[0.1em]
    Radius & 0 & 1 & \cellcolor{lightgray!50}2
    \\\midrule
    $\alpha=0.01$ & 56.18 & 58.23 & \cellcolor{lightgray!50}58.75 \\
    $\alpha=0.03$ & 94.97 & 95.07 & \cellcolor{lightgray!50}95.08 \\
    $\alpha=0.10$ & 98.71 & 98.72 & \cellcolor{lightgray!50}99.93 \\
    \bottomrule[0.1em]
     \end{tabular}}
}
\hfill
\vspace{-0.5cm}
\end{table*}

\noindent\textbf{Bechmarking State-of-the-art Methods.}
There exist several categories of approaches that could handle this problem.
The direct rival to our method is training models on synthetic datasets and testing on real data, including DISCNet~\cite{feng2021removing} and BNUDC~\cite{koh2022bnudc}.
Besides the officially released pre-trained models, we also include these networks trained with our dataset and training strategy.
Image-to-image translations offer another avenue to solve this task.
It remains challenging to build mappings between complex domains using existing style transfer algorithms, and they are usually dominated by global image distribution and overlook the detailed local structures (\eg, flare, blur, noise).
We include two representative image translation methods, MUNIT~\cite{huang2018munit} and TSIT~\cite{jiang2020tsit} for comparison.

Table~\ref{tab:sota_comparison} summarizes the benchmark.
Pre-trained models (\ie, DISCNet and BNUDC) achieve relatively low PSNR, SSIM and LPIPS values compared to \textit{Aligned Ref}, and it shows poorer performance on CD and SIFID, indicating domain discrepancy against the reference images.
In contrast, the same models retrained on our dataset (5th and 6th row) demonstrate noticeable improvement over these two sets of metrics.
Image translation approaches only capture statistics over the distribution of the whole dataset, and thus neglect local details, leading to a performance drop on the test set.
The PPM-UNet trained with image pairs generated from AlignFormer achieves comparable or the best performance on all sets of measurements.
These results show the effectiveness of our AlignFormer and PPM-UNet.

Figure~\ref{fig:sota_comparison} qualitatively compares the benchmarks in Table~\ref{tab:sota_comparison}.
All pre-models fail to suppress flare and blur, and they cannot restore plausible color and contrast.
On the other hand, the models trained with our datasets show better-restored results.
The results of ``AlignFormer+DISCNet'' and ``AlignFormer+BNUDC'' accurately remove flare.
However, the result of ``AlignFormer+BNUDC'' suffers from regional inconsistency (See yellow arrow).
Image translation approaches cannot accurately recover the light source regions, and produce considerable noise and artifacts.
Among these benchmarks, our results have sharper details with a similar domain style.

\vspace{-0.2cm}
\subsection{Ablation Study}
\begin{figure}[t]
    \centering
    \includegraphics[width=.9\linewidth]{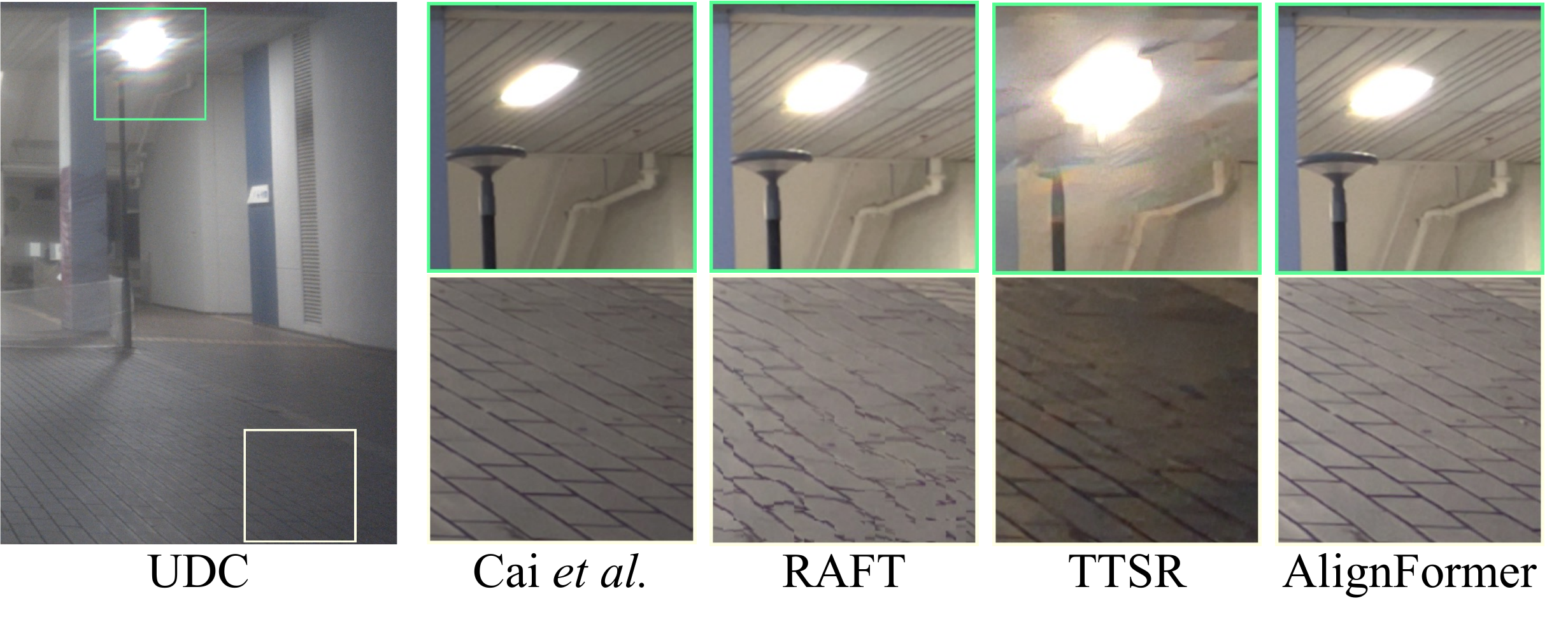}
    \vspace{-0.4cm}
    \caption{\textbf{Ablation study on alignment method.}
}
    \label{fig:alignment}
    \vspace{-0.6cm}
\end{figure}

\vspace{-0.1cm}
\noindent\textbf{Effectiveness of Alignment Method.}
As there is no ground-truth correspondence, 
it is non-trivial to quantify how well the pseudo GT is aligned to the UDC image, especially when severe domain inconsistency exists.
Thus, we indirectly measure the displacement error with LoFTR~\cite{sun2021loftr} serving as a keypoint matcher.
Given a set of matched keypoints for both images, PCK measures the percentage of correct keypoints transferred to another image, which lie within a certain radius of the same coordinates (ideally $0$ if two images are well-aligned). Please refer to the supplementary material for details.
Suppose $d$ is the displacement of a pair of matched keypoint,
the keypoint pair is correctly aligned when $d<\alpha\times\mathrm{max}(H, W)$, where $\alpha$ is the threshold and $H$, $W$ represent the height and width of the image.
Table~\ref{tab:alignment} shows the accuracy of alignment among warping-based methods (RAFT~\cite{teed2020raft}, Cai \etal~\cite{cai2019toward}).
As presented, our AlignFormer more accurately align to the corresponding UDC images than the compared methods.

Besides visualizing the results of warping-based methods, we also include the result of TTSR~\cite{yang2020learning} as reference in Figure~\ref{fig:alignment}. 
Flow-based alignment yields distorted structures, while Cai \etal~\cite{cai2019toward} introduces illuminance change and blur.
TTSR cannot recover high-quality details around the light region.
Our method can alleviate spatial misalignment well.


\noindent\textbf{Optical Flow Guidance.}
We incorporate three pre-trained optical flow estimators: SPyNet \cite{ranjan2017optical}, PWC-Net \cite{sun2018pwc}, and RAFT \cite{teed2020raft} into our AlignFormer. The results in Table~\ref{tab:exp_optical} show that the performance can be improved with better optical flow estimation.
The model with higher accuracy (RAFT) exhibits less displacement on the resulting image.
Removing flow guidance, denoted as \textit{Zero}, remarkably deteriorates the performance.

\noindent\textbf{Radius of Local Grid.}
As shown in Table~\ref{tab:exp_radius}, we change the radius $r$, which specifies the size of the local grid used to search neighboring samples in the flow-guided sampler. When the radius is $0$, the feature is retrieved at one single point given by the flow offset, which can be regarded as a warping strategy in the feature space. It can be observed that even with a radius $0$, we can still get rough information from the optical flow.
\noindent\textbf{Effectiveness of DAM.}
Table~\ref{tab:exp_dam} shows that the domain alignment module drives more accurate alignment for subsequent GAM and facilitates more robust attention compared to that without DAM.

\begin{table}[t]
 \centering
 \caption{\textbf{Ablation experiments on DAM and PPM.}
 }
\vspace{-0.4cm}
\subfloat[Ablation study on DAM.]{
    \label{tab:exp_dam}
    \hfill
    \resizebox{.49\linewidth}{!}{
     \begin{tabular}{c|cc}
   \toprule[0.1em]
    PCK (\%)$\uparrow$ & w/o DAM & \cellcolor{lightgray!50}w/ DAM
    \\\midrule
    $\alpha=0.01$ & 56.15 & \cellcolor{lightgray!50}58.75 \\
    $\alpha=0.03$ & 94.70 & \cellcolor{lightgray!50}95.08 \\
    $\alpha=0.10$ & 98.66 & \cellcolor{lightgray!50}99.93 \\
    \bottomrule[0.1em]
     \end{tabular}}
}
\hfill
\subfloat[Ablation study on PPM.]{
    \label{tab:exp_ppm}
    \hfill
    \resizebox{.44\linewidth}{!}{
     \begin{tabular}{c|cc}
   \toprule[0.1em]
     & w/o PPM & \cellcolor{lightgray!50}w/ PPM
    \\\midrule
    PSNR$\uparrow$ & 22.68 & \cellcolor{lightgray!50}22.95\\
    SSIM$\uparrow$ & 0.8670 & \cellcolor{lightgray!50}0.8581\\
    LPIPS$\downarrow$ & 0.1328 &  \cellcolor{lightgray!50}0.1236\\
    \bottomrule[0.1em]
     \end{tabular}}
}\hfill
\vspace{-0.5cm}
\end{table}
\noindent\textbf{Effectiveness of PPM.}
The PPM layers propagate global prior into the network and stabilize training and suppress artifacts in UDC image restoration. In Table~\ref{tab:exp_ppm}, we remove Pyramid Pooling Module (w/o PPM) and the performance drops noticeably.

\vspace{-0.3cm}
\section{Conclusions}
We have presented AlignFormer for generating high-quality and well-aligned pseudo UDC pairs. The key insight of our pipeline is to exploit the pixel correspondence by both semantic and geometric cues embedded into a Transformer-based structure.  
Apart from the novel designs, we also contribute a new dataset to support UDC image restoration networks training, a step towards solving UDC image restoration in the real world.
%

\noindent\textbf{Acknowledgement.}
\footnotesize{This study is supported under the RIE2020 Industry Alignment Fund Industry Collaboration Projects (IAF-ICP) Funding Initiative, a Direct Grant from CUHK, the National Key R\&D Program of China (NO.2022ZD0160100), as well as cash and in-kind contribution from the industry partner(s).}

{\small
\bibliographystyle{ieee_fullname}
\bibliography{egbib}
}

\end{document}


\title{Generating Aligned Pseudo-Supervision from Non-Aligned Data for \\ Image Restoration in Under-Display Camera\\
- Supplementary Material -}

\author{
Ruicheng Feng$^{1}$\quad
Chongyi Li$^{1}$\quad
Huaijin Chen$^{2}$\quad
Shuai Li$^{2}$\quad
Jinwei Gu$^{3,4}$\quad
Chen Change Loy$^{1}$
\\
$^{1}$S-Lab, Nanyang Technological University\quad
$^{2}$SenseBrain Technology\\
$^{3}$The Chinese University of Hong Kong\quad
$^{4}$Shanghai AI Laboratory\\
{\tt\small \{ruicheng002, chongyi.li, ccloy\}@ntu.edu.sg}
\\
{\tt\small \{huaijin.chen, shuailizju\}@gmail.com\quad
jwgu@cuhk.edu.hk}
}

\maketitle

\appendix
\setcounter{table}{0}
\renewcommand{\thetable}{A\arabic{table}}
\setcounter{figure}{0}
\renewcommand{\thefigure}{A\arabic{figure}}

\section{Implementation Details}
\subsection{Camera Setup}
\begin{figure}[t]
    \centering
     \includegraphics[width=0.9\linewidth]{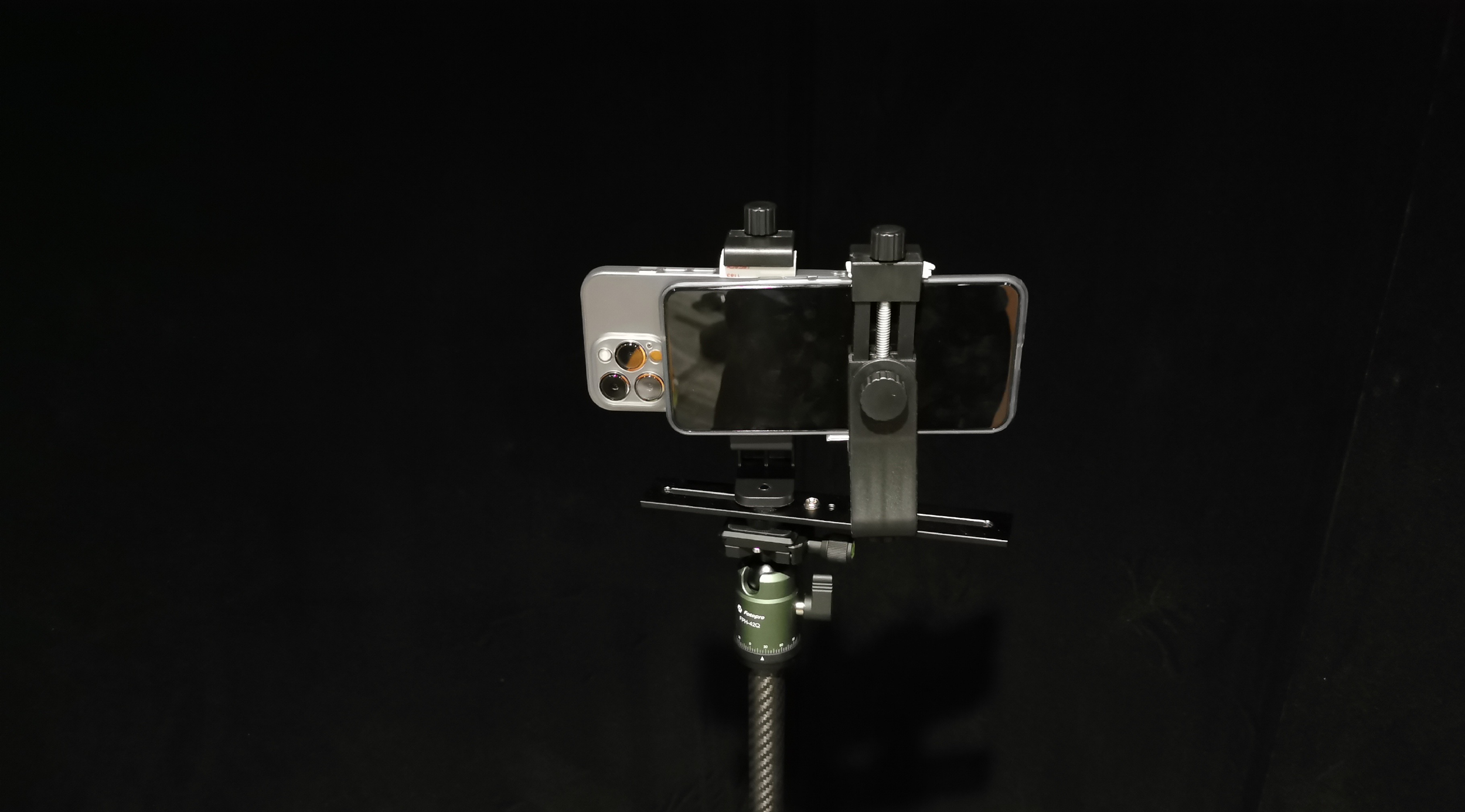}
    \caption{A close-up picture of our custom-built camera setup consisting of two smartphones mounted on a tripod.}
    \label{fig:setup}
\end{figure}

\begin{figure*}[t]
    \centering
    \includegraphics[width=.96\linewidth]{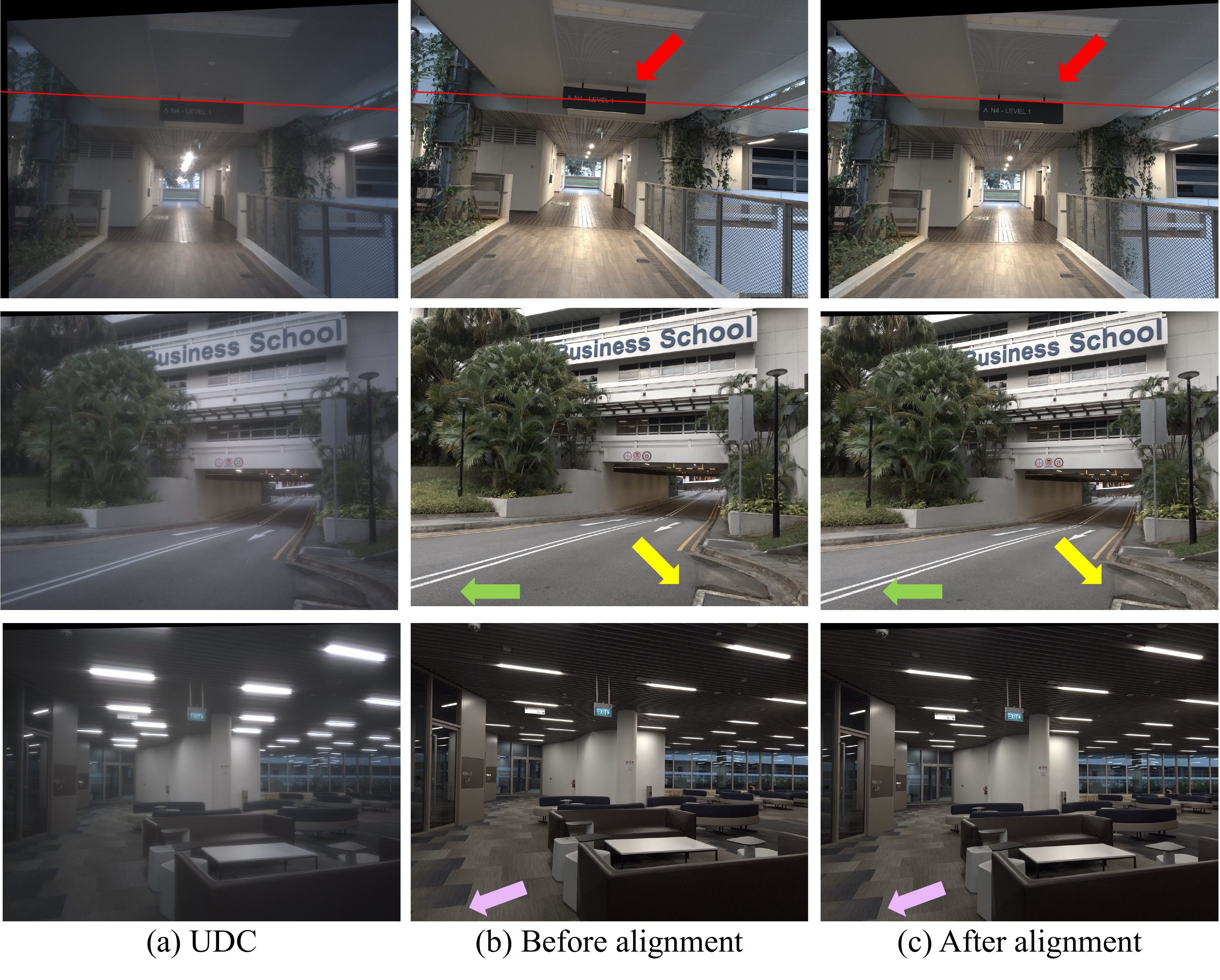}
    \vspace{-0.4cm}
    \caption{\textbf{The image examples in our dataset.}
    Images (a) and (b) are captured by UDC and a normal camera, and they exhibit obvious misalignment. We roughly align (b) to (a) with homography transformation to obtain (c).
    Even after alignment, there still exists mild misalignment between (a) and (c), which will be mitigated by our AlignFormer.
    }
    \label{fig:triplet_samples}
    \vspace{-0.2cm}
\end{figure*}

To build a real-world dataset, we mount the two smartphones on the tripod as shown in Figure \ref{fig:setup} and take each set of images using Bluetooth remote controller to control the shutter speed.
The camera modules are physically placed as close as possible to decrease the baseline in a stereo setting.
The aperture and focal length are fixed and unadjustable for both cameras.
The resolution of ZTE Axon 20 is $3264\times2448$, and that of iPhone 13 Pro is $4032\times3024$.
For the \textit{degraded image}, we set the under-display camera configurations by its built-in automatic exposure system and take three shots bracketed at $[1, 1/4, 1/16]$, which are then composed into one HDR image.
For the \textit{reference image}, we set a low ISO value ranging from $100$ to $200$ to avoid heavy noise, and adjust the shutter speed to capture sharp and clean images of proper exposure.
We avoid capturing objects that are too close to filter out image pairs with large parallax or occlusion.
For each scene of the pair, we register the image captured by iPhone as a reference image to the UDC image using a homography transform calculated by RANSAC \cite{vedaldi2010vlfeat}.
Then we crop out the invalid areas due to the homography transformation and parallax between two cameras, and downsample the pairs to  $3200\times 2400$.
The dataset is exemplified by triplet sets in Figure \ref{fig:triplet_samples}. The reference images after alignment still show mild displacement compared to the corresponding UDC images.


\subsection{Occlusion Mask}
Occluded pixels, by definition, are invisible in the reference image, which should be discounted for spatial supervision (\eg, $\mathcal{L}_1$ and $\mathcal{L}_{VGG}$), since inaccurate deformations over these regions could deteriorate image restoration network training.
We implement the detection based on forward-backward consistency assumption~\cite{sundaram2010dense}, that is, for non-occluded pixels, traversing the forward flow and then backward should arrive at the same pixel.

\begin{figure}[t]
    \centering
    \includegraphics[width=\linewidth]{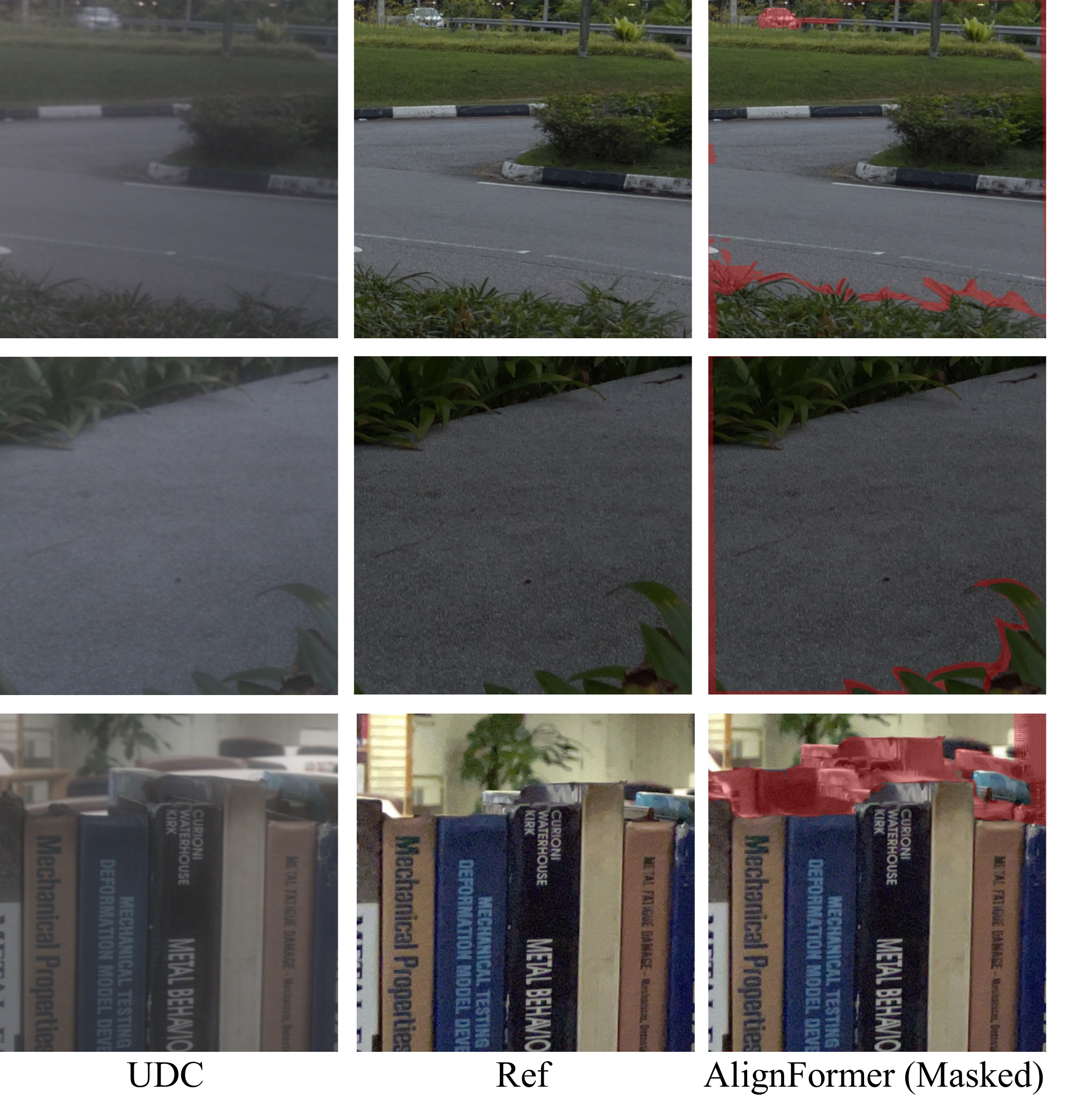}
    \vspace{-0.3cm}
    \caption{\textbf{Invalid mask visualization.}
    The invalid (occlusion) mask (highlighted in red) is estimated by forward-backward assumption. Note that it also detects pixels at borders, and moving objects in non-still images.
    }
    \label{fig:occlusion_mask}
    \vspace{-0.3cm}
\end{figure}

In particular, the forward optical flow from $I_R$ to $\hat{I_D}$ is given by $\Psi^{f}=\psi_{flow}(\hat{I_D}, I_R)$, and the backward flow can similarly be estimated by $\Psi^{b}=\psi_{flow}(I_R, \hat{I_D})$.
Suppose forward flow vector is $\mathbf{w}=\Psi^{f}(\mathbf{p})$ at point $\mathbf{p}=(x,y)$, the forward-backward flow vector is denoted as $\hat{\mathbf{w}}=\Psi^{b}(\mathbf{p}+\Psi^{f}(\mathbf{p}))$.
The non-occlusion mask $M$, with granted tolerance for small estimation errors, can then be formulated by
\begin{equation}
    M(\vb{p})=
    \begin{cases}
        1, &\text{if $\|\mathbf{w}+\hat{\mathbf{w}}\|_2<\alpha(\|\mathbf{w}\|_2+\|\hat{\mathbf{w}}\|_2)+\beta$}\\
        0, &\text{otherwise}
    \end{cases},
\end{equation}
where tolerance parameters $\alpha$ and $\beta$ are set to $0.1$ and $1$ in this paper.
Figure~\ref{fig:occlusion_mask} illustrates several examples of masked invalid regions (highlighted in red).

\subsection{Conditional PatchGAN}
To further improve the visual quality, we also add adversarial loss based on conditional PatchGAN~\cite{isola2017image}:
\begin{equation}
    \mathcal{L}_{GAN}=-\mathbb{E}[\text{log}\mathcal{D}(I_D, I_O)],
\end{equation}
where $\mathcal{D}$ is the discriminator conditioned on both input and output of PPM-UNet, and it is optimized by
\begin{equation}
    \mathcal{L}_{\mathcal{D}}=-\mathbb{E}[\text{log}\mathcal{D}(I_D, I_P)]
    -\mathbb{E}[\text{log}(1-\mathcal{D}(I_D, I_O))].
\end{equation}

PatchGAN uses a discriminator that distinguishes image patches of size $70\times 70$, which proves to produce sharper details than the vanilla ``global'' discriminator.
Inspired by \cite{kupyn2018deblurgan,kupyn2019deblurgan}, we adopt conditional PatchGAN as a discriminator to capture high frequencies in local features.
Particularly, we use the discriminator architecture that only penalizes structure at the scale of patches (PatchGAN). This discriminator tries to classify if each $N\times N$ patch is real or fake. We run the discriminator across the image, and then average all responses to obtain the patch-based output.
$N$ is set to $16$ in our experiments.
We also empirically found conditional GAN, where the discriminator is conditioned on UDC images, facilitates more realistic results.

\begin{figure}[t]
    \centering
    \includegraphics[width=\linewidth]{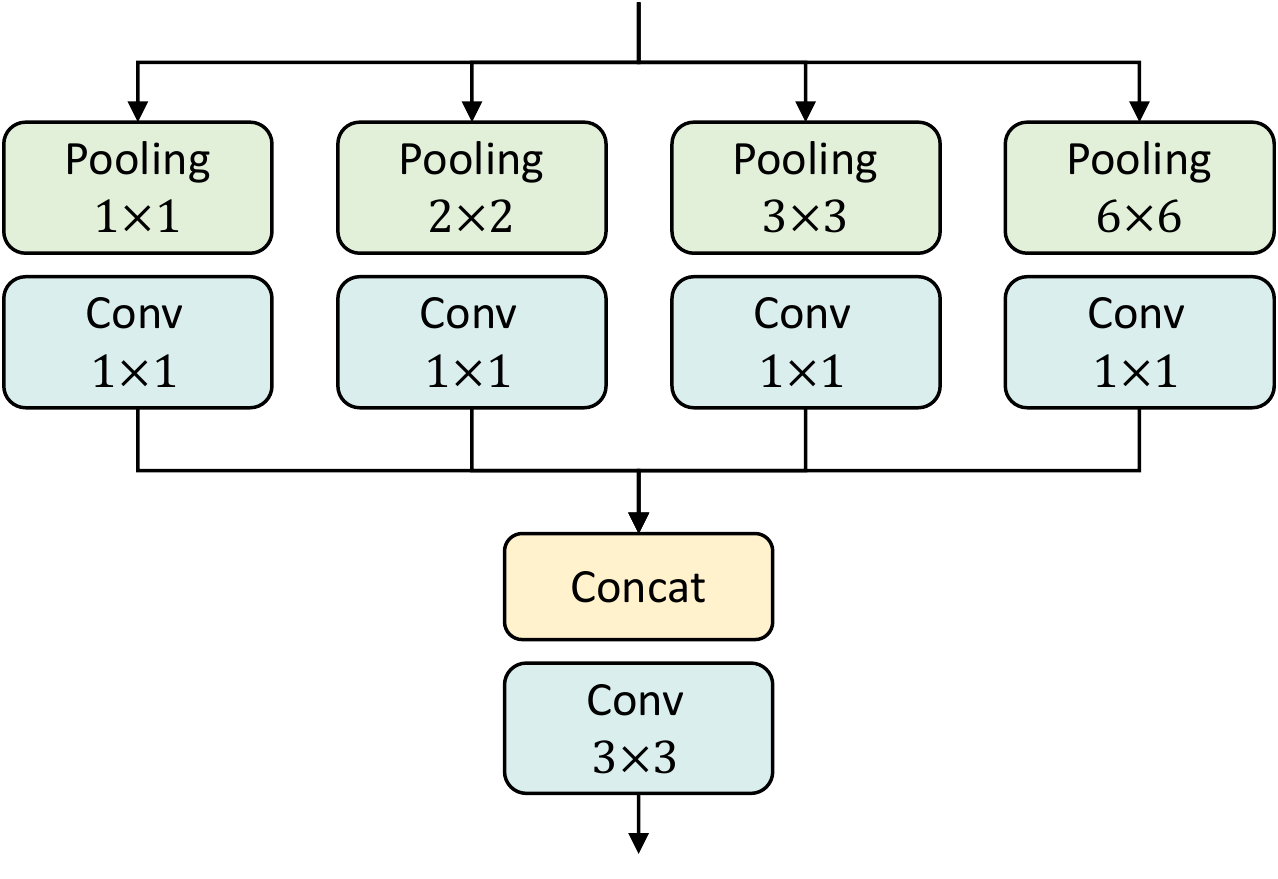}
    \vspace{-0.3cm}
    \caption{\textbf{Illustration of Pyramid Pooling Module (PPM).}
    }
    \label{fig:ppm_arch}
    \vspace{-0.3cm}
\end{figure}
\subsection{Network Structure}
The Domain Alignment Module (DAM) contains two sub-nets: a guidance net and a matching net.
The detailed architecture of DAM is listed in Table~\ref{tab:structure_dam}. In the matching net, StyleConv is the conv layer modulated by style conditions as proposed in StyleGANv2~\cite{karras2020analyzing}.
The guidance net is designed to generate a conditional vector that extracts holistic domain information from the reference image.
After that, the matching net transfers the domain information, \eg, color, illuminance, and contrast, to the degraded UDC image and produces a coarse restored image that is similar to the reference image.

The structure of the image restoration network is shown in Table~\ref{tab:structure_restoration}. We adopt a modified U-Net as in \cite{zhou2022lednet} and add a Pyramid Pooling Module (PPM) \cite{zhao2017pyramid} into the network to capture global information (See Figure~\ref{fig:ppm_arch}).
We adopt the original design of PPM containing $4$ mean pooling branches with bin sizes of {1, 2, 3, 6}. As demonstrated in the main paper, the PPM layers propagate the global prior into the network and stabilize training and suppress artifacts in UDC image restoration.

\begin{table}[t]
\caption{\textbf{Detailed structure of DAM.}
    $k$ and $s$ indicate the kernel size and stride of the convolutional layer.
    $\uparrow$ and $\downarrow$ represent $2\times$ upsampling and $2\times$ downsampling, respectively.
}
\label{tab:structure_dam}
    \centering
    \resizebox{.95\linewidth}{!}{
    \begin{tabular}{@{\hskip 0.1in}l@{\hskip 0.1in}cl}
    \toprule
    \multicolumn{3}{c}{Guidance Net}\\
    Layer & Configuration & Output size\\
    \midrule
    Conv, LeakyReLU & $k=3,s=1$ & $256\times 256\times 64$\\
    Conv, LeakyReLU & $k=3,s=2$ & $128\times 128\times 64$\\
    Conv, LeakyReLU & $k=3,s=1$ & $128\times 128\times 64$\\
    Conv, LeakyReLU & $k=3,s=1$ & $64\times 64\times 64$\\
    Conv            & $k=3,s=1$ & $64\times 64\times 64$\\
    Global Average Pooling & - & $1\times 1\times 64$\\
    \bottomrule
    \toprule
    \multicolumn{3}{c}{Matching Net}\\
    Layer & Configuration & Output size\\
    \midrule
    StyleConv, LeakyReLU & $k=3,s=1$ & $256\times 256\times 64$\\
    StyleConv, LeakyReLU, $\downarrow$ & $k=3,s=1$ & $128\times 128\times 64$\\
    StyleConv, LeakyReLU & $k=3,s=1$ & $128\times 128\times 64$\\
    StyleConv, LeakyReLU, $\downarrow$ & $k=3,s=1$ & $64\times 64\times 64$\\
    StyleConv, LeakyReLU & $k=3,s=1$ & $64\times 64\times 64$\\
    StyleConv, LeakyReLU & $k=3,s=1$ & $64\times 64\times 64$\\
    StyleConv, LeakyReLU & $k=3,s=1$ & $64\times 64\times 64$\\

    StyleConv, LeakyReLU, $\uparrow$ & $k=3,s=1$ & $128\times 128\times 64$\\
    StyleConv, LeakyReLU & $k=3,s=1$ & $128\times 128\times 64$\\
    StyleConv, LeakyReLU, $\uparrow$ & $k=3,s=1$ & $256\times 256\times 64$\\
    StyleConv, LeakyReLU & $k=3,s=1$ & $256\times 256\times 64$\\
    StyleConv, LeakyReLU & $k=3,s=1$ & $256\times 256\times 3$\\

    \bottomrule
    \end{tabular}
    }
\end{table}

\begin{table}[t]
\caption{\textbf{The MTF curve results.}
Reported are the weighted mean summary of several detected slant edges.
}
\label{tab:exp_mtf}
\centering
\resizebox{.9\linewidth}{!}{
\begin{tabular}{l|ccc}
\toprule
Metric & UDC & Ref & Restored
\\\midrule
MTF50 (LW/PH) $\uparrow$ & 661 & 1516 & 1039 \\
MTF20 (LW/PH) $\uparrow$ & 1307 & 2023 & 1769 \\
\bottomrule
\end{tabular}
}
\end{table}

\begin{table*}[t]
\caption{\textbf{Detailed structure of PPM-UNet.}}
\label{tab:structure_restoration}
    \centering
    \resizebox{.95\linewidth}{!}{
    \begin{tabular}{@{\hskip 0.1in}c@{\hskip 0.1in}cccccc}
    \toprule
    Module & Kernel size & \# of channels & Dilation & Stride & Activation & Output size\\
    \midrule
    Conv1  & $3\times3$ & 32 & 1 & 1 & LeakyReLU ($0.2$) & $256\times 256\times 32$\\
    Conv2 & $3\times3$ & 64 & 1 & 2 & LeakyReLU ($0.2$) & $128\times 128\times 64$\\
    Conv3 & $3\times3$ & 64 & 1 & 1 & LeakyReLU ($0.2$) & $128\times 128\times 64$\\
    PPM1 & - & - & - & - & - & $128\times 128\times 64$\\
    Conv4 & $3\times3$ & 128 & 1 & 2 & LeakyReLU ($0.2$) & $64\times 64\times 128$\\
    Conv5 & $3\times3$ & 128 & 1 & 1 & LeakyReLU ($0.2$) & $64\times 64\times 128$\\
    PPM2 & - & - & - & - & - & $64\times 64\times 128$\\
    Conv6 & $3\times3$ & 128 & 1 & 2 & LeakyReLU ($0.2$) & $32\times 32\times 128$\\
    Conv7 & $3\times3$ & 128 & 1 & 1 & LeakyReLU ($0.2$) & $32\times 32\times 128$\\
    PPM3 & - & 128 & - & - & - & $32\times 32\times 128$\\
    Conv8 & $3\times3$ & 128 & 1 & 1 & LeakyReLU ($0.2$) & $32\times 32\times 128$\\
    Conv9 & $3\times3$ & 128 & 1 & 1 & LeakyReLU ($0.2$) & $32\times 32\times 128$\\
    \midrule
    Add (w/ PPM3) & - & 128 & - & - & - & $32\times 32\times 128$\\
    Upsample $\uparrow$ & - & 128 & 2 & - & - & $64\times 64\times 128$\\
    Conv10 & $3\times3$ & 128 & 1 & 1 & LeakyReLU ($0.2$) & $64\times 64\times 128$\\
    Conv11 & $3\times3$ & 128 & 1 & 1 & LeakyReLU ($0.2$) & $64\times 64\times 128$\\
    \midrule
    Add (w/ PPM2) & - & 128 & - & - & - & $64\times 64\times 128$\\
    Upsample $\uparrow$ & - & 128 & 2 & - & - & $128\times 128\times 128$\\
    Conv12 & $3\times3$ & 64 & 1 & 1 & LeakyReLU ($0.2$) & $128\times 128\times 64$\\
    Conv13 & $3\times3$ & 64 & 1 & 1 & LeakyReLU ($0.2$) & $128\times 128\times 64$\\
    \midrule
    Add (w/ PPM1) & - & 64 & - & - & - & $128\times 128\times 64$\\
    Upsample $\uparrow$ & - & 64 & 2 & - & - & $256\times 256\times 64$\\
    Conv14 & $3\times3$ & 32 & 1 & 1 & LeakyReLU ($0.2$) & $256\times 256\times 32$\\
    Conv15 & $3\times3$ & 32 & 1 & 1 & LeakyReLU ($0.2$) & $256\times 256\times 32$\\
    \midrule
    Conv16 & $3\times3$ & 3 & 1 & 1 & - & $256\times 256\times 3$\\
    \bottomrule
    \end{tabular}
}
\end{table*}

\section{Objective Evaluation}
\begin{figure}[t]
    \centering
    \includegraphics[width=.9\linewidth]{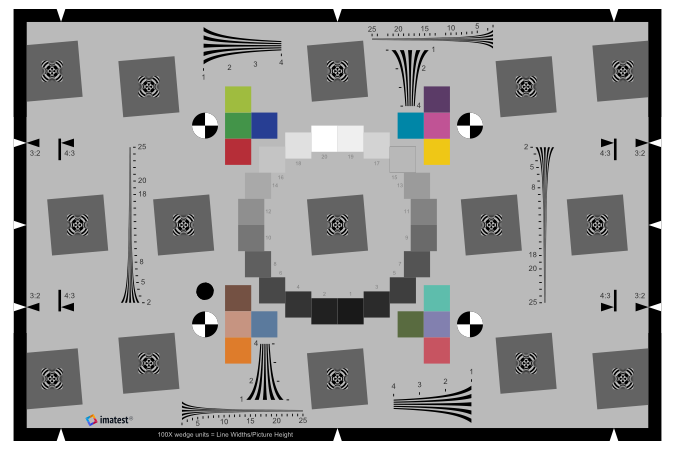}
    \vspace{-0.2cm}
    \caption{Lab-based image of ISO 12233 eSFR test chart.
    }
    \label{fig:iso}
    \vspace{-0.2cm}
\end{figure}
\begin{figure*}[t]
    \centering
    \includegraphics[width=.98\linewidth]{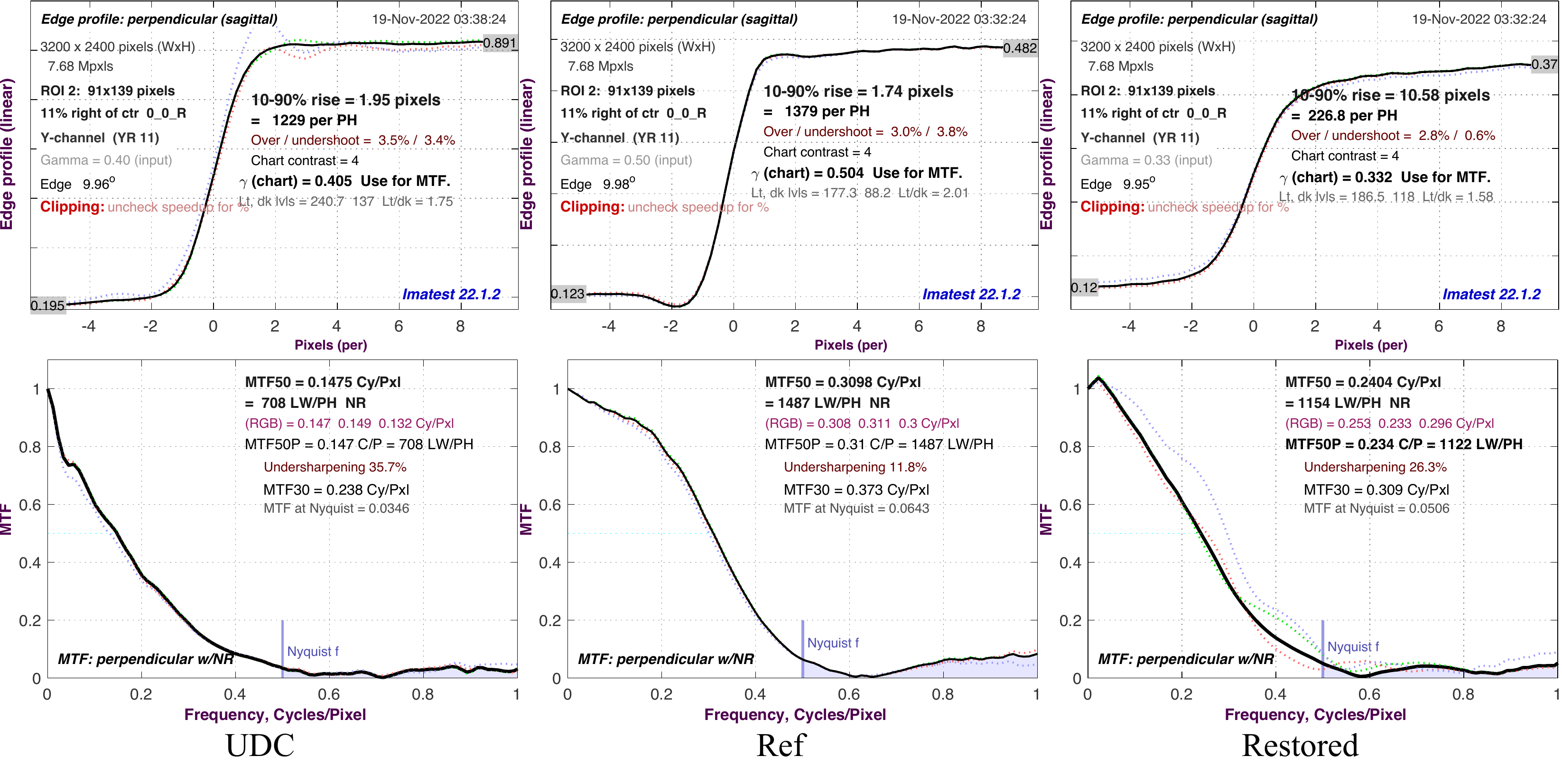}
    \vspace{-0.3cm}
    \caption{MTF curves and edge profiles,
    obtained with Imatest on ISO12233 test chart.
    }
    \label{fig:mtf}
\end{figure*}

In addition to quantitatively measuring the characteristics of the captured natural scenes (\ie, on the test dataset), we also conduct objective quality evaluation via engineering-based quality metrics.
The Modulation Transfer Function (MTF) and the related Spatial Frequency Response (SFR) are commonly used to characterize an imaging system’s reproduction of modulation, as a function of spatial frequency~\cite{van2019edge,vint2019evaluation, rego2022deep}.
MTF can inform the system's resolution and sharpness, which mainly contribute to the overall image quality.
MTF can be derived from the slanted-edge technique \cite{burns2000slanted} with carefully designed test charts under strict laboratory conditions.
To calculate the MTF curve, we use the Enhanced version of ISO 12233 imatest eSFR test chart~\cite{iso12233} (See Figure~\ref{fig:iso}).
The enhanced eSFR ISO test chart adds 6 squares on sides, 16 color patches, and several wedge patterns.

A key data point from the MTF curve is MTF50, where MTF is 50\% of its low (0) frequency value. Similarly, MTF20 is the spatial frequency where MTF is 20\% of the zero frequency.
As recommended by Imatest~\cite{imatest}, we use these two metrics for analysis throughout this work.
We use the data dumps with least post-processing, such that the evaluation can operate in a more linear region, and hence results are less affected by overexposure, underexposure, and excessive sharpening.
Table~\ref{tab:exp_mtf} summarizes the results.
As can be observed, the MTF values increase when the UDC images are restored by our PPM-UNet, which demonstrates that the image restoration network also increases the contrast and sharpness of images.

In addition, we show the complete MTF curves, and the edge profile the MTF is derived from, of the original UDC, reference, and restored UDC in Figure~\ref{fig:mtf}. One can observe the contrast improvement in the low- to mid-frequency band of the restored image compared to the original UDC image, as the modulation transfer is noticeably higher in the 0-0.4 cycles/pixel frequency region. The overall MTF shape of the restored image is also more similar to the reference image compared to the original UDC, suggesting an overall more natural contrast and sharpness after restoration.

\section{Analysis of Displacement Metrics}
\begin{figure*}[t]
    \centering
    \includegraphics[width=.96\linewidth]{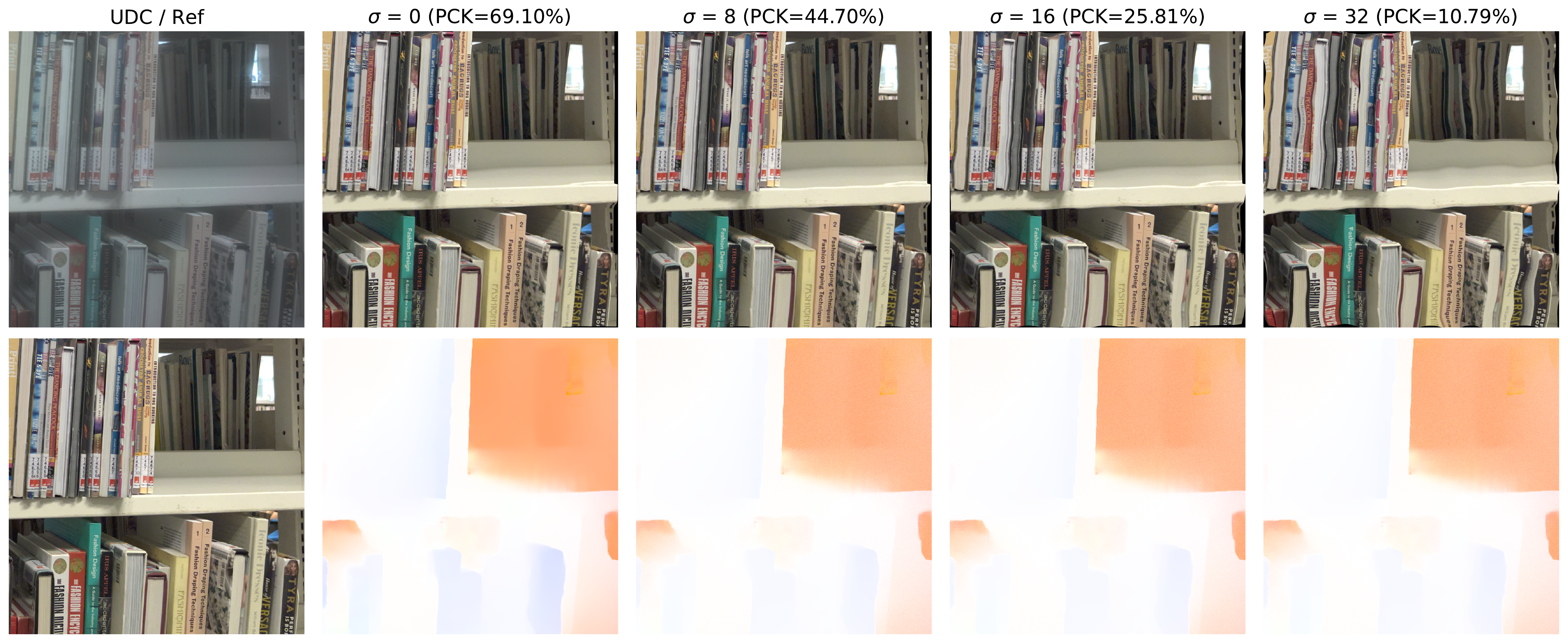}
    \caption{\textbf{Illustration of PCK and flows with perturbation.}
    Top row are the image and wrapped images, while bottom row show the flows with perturbation.
    The wrapped images are severely distorted when injecting large perturbation on flow.
    Reported results are evaluated on PCK with $\alpha=0.01$ ($\sim$ 1 pixel error tolerance).
    }
    \label{fig:pck_flow}
\end{figure*}
\begin{figure}[t]
    \centering
    \includegraphics[width=.96\linewidth]{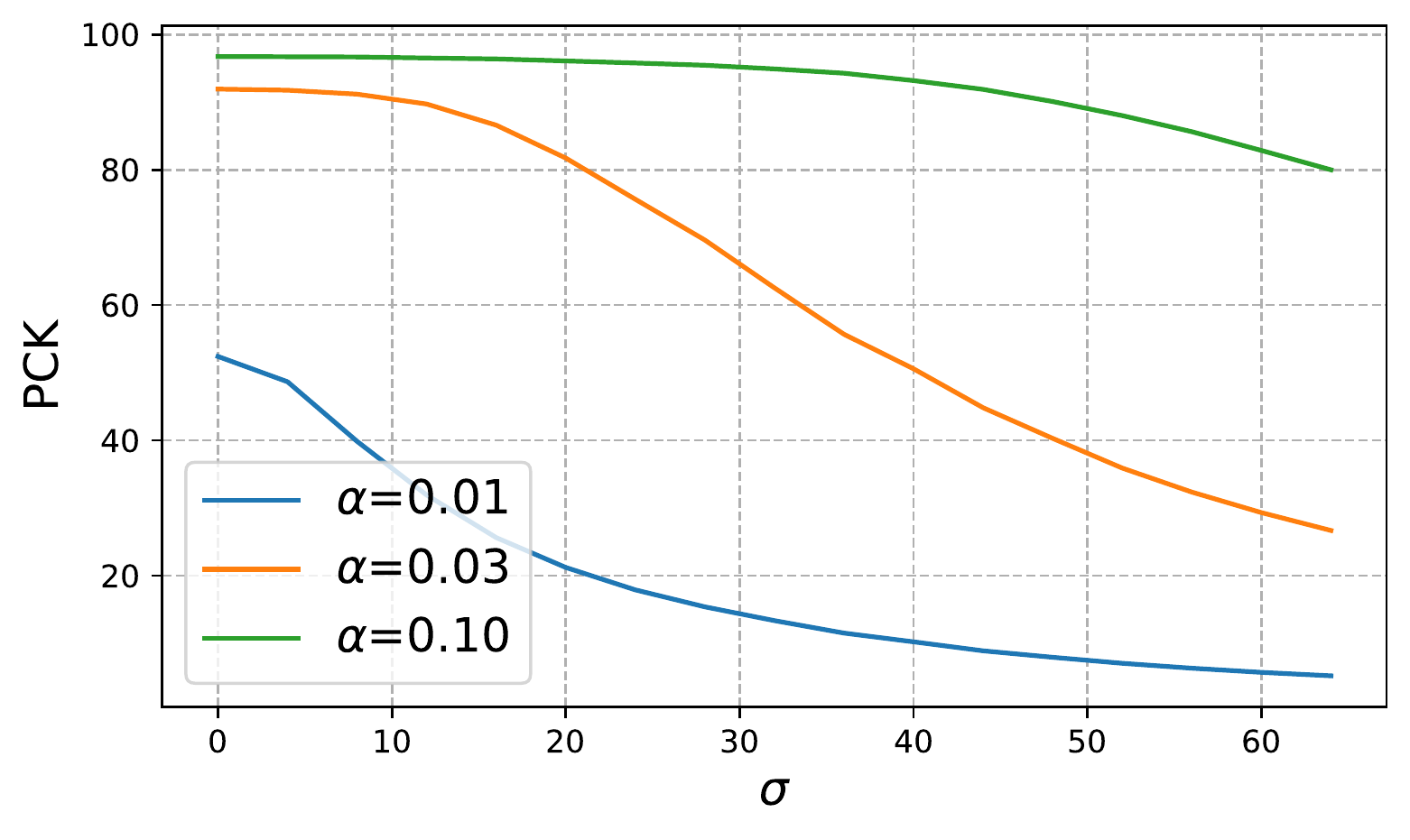}
    \caption{\textbf{PCK on different wrapped images where flows are added with various perturbations.}
    The perturbation is controlled by Gaussian noise with various std $\sigma$ ranging from $0$ (Original flow) to $64$. We report the PCK results averaged  over 330 images of shape $1024\times 1024$.
    The PCK curves exhibit a consistent tendency with perturbation, indicating the displacement metric PCK can accurately reflect misalignment in our task.
    }
    \label{fig:pck_std}
\end{figure}
In absence of ground-truth correspondence, it is non-trivial to quantify how well the pseudo GT (output of AlignFormer) is aligned to the UDC image.
Thus, we indirectly measure the displacement error with LoFTR~\cite{sun2021loftr} that serves as a keypoint matcher.
Given a set of matched keypoints from two images, PCK measures the Percentage of Correct Keypoints transferred to another image, which lie within a certain error threshold.

In particular, suppose $\bm{x}_A$ and $\bm{x}_B$ are the same matched keypoint located at $\mathbf{p}_A$ in one image and $\mathbf{p}_B$ in another image, $d=\|\mathbf{p}_A-\mathbf{p}_B\|_2$ measures the displacement of coordinates.
Ideally, the offset should be all zeros when two images are perfectly aligned.
As global keypoint search may lead to long-range matching, causing outliers with large displacement, we do not average out all displacement errors $d$ over detected keypoints.
Instead, we calculate the percentage of correct matched pairs, denoted as \textit{PCK} in the main paper.
The keypoint pair is deemed correctly aligned when $d$ is smaller than the preset threshold.
Following common practice in image matching \cite{peebles2022gansupervised,jiang2021cotr}, we set $\alpha$ error threshold, given by $d<\alpha\times \text{max}(H,W)$, where $H$ and $W$ are the height and width of the image.

To justify the metrics, we conduct controlled experiments for further analysis. 
Specifically, we estimate optical flow from the reference image to the UDC image using RAFT~\cite{teed2020raft}, inject perturbation into the flow, and then warp the reference image with the deformed flow.
Following~\cite{yang2020deep}, the deformation is implemented by sampling amounts of independent Gaussian noise with $\mu=0$ and controllable standard deviation (std) $\sigma$.
Then we compute PCK between the warped reference image and UDC image.
The reported PCK results are averaged over $330$ images of size $1024\times 1024$.
Generally, a larger perturbation to the flow would induce greater displacement on the warped image.
Figure~\ref{fig:pck_std} reflects the behaviors of PCK at various $\alpha$ thresholds under different perturbations. All curves consistently drop when the perturbations (controlled by $\sigma$) become greater.
Figure~\ref{fig:pck_flow} visualizes an example of warping flow with perturbation. We also observe a similar tendency of PCK and perturbation.
This implies the PCK metric is suitable in our cases for quantifying misalignment.

\section{Additional Ablation Studies}
\begin{figure}[t]
    \centering
     \includegraphics[width=0.9\linewidth]{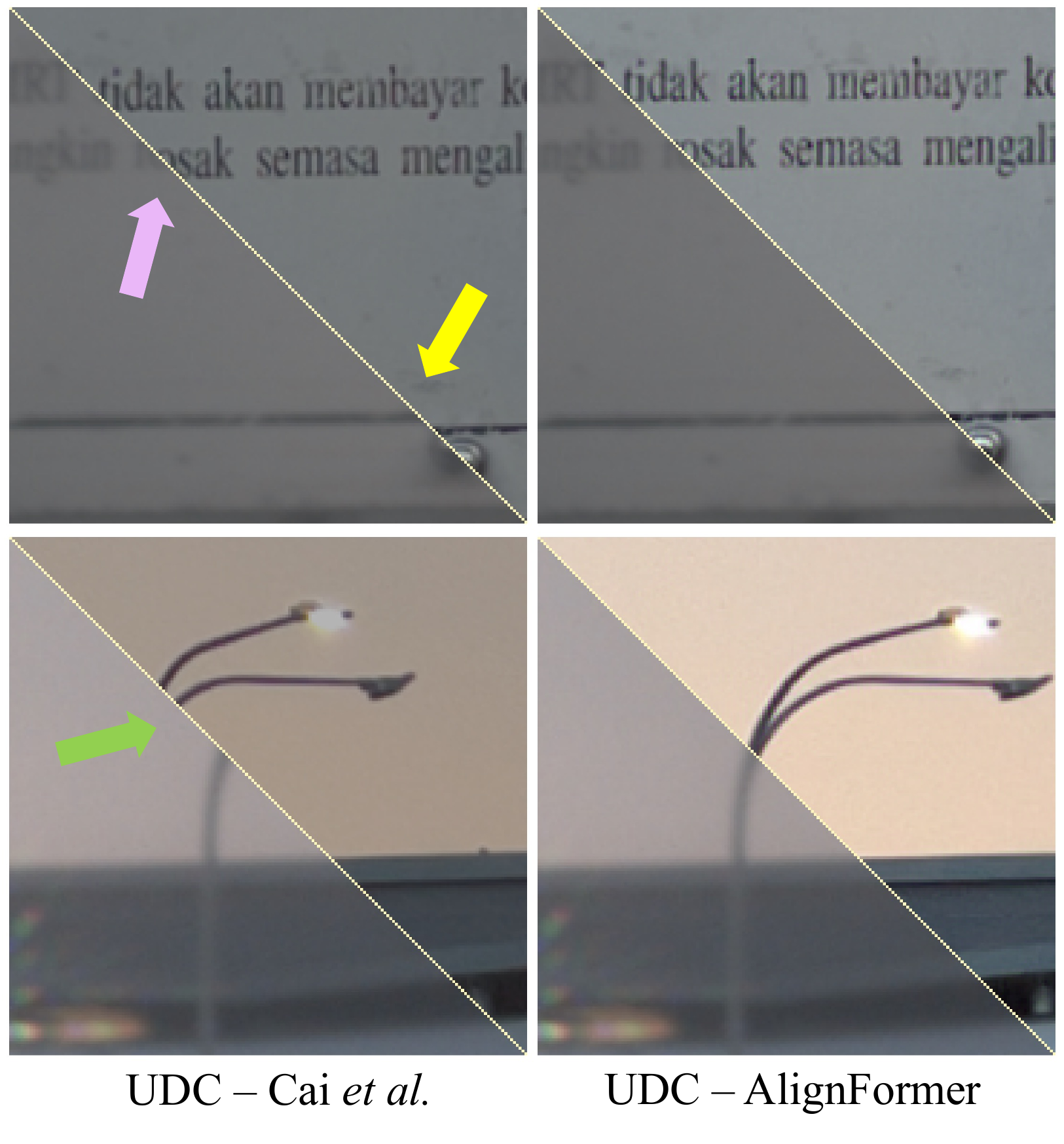}
     \caption{\textbf{Stitched pairs of local patches from aligned images.}
     Bottom left are UDC images, and top right are corresponding results generated by Cai~\etal~\cite{cai2019toward} or AlignFormer.
     Results demonstrate Cai~\etal~\cite{cai2019toward} cannot achieve accurate registration, while our AlignFormer perferctly aligns the reference images.}
    \label{fig:stitch_pairs}
\end{figure}

\paragraph{Effectiveness of Alignment Method.}
Figure 7 in the main body presents the results aligned by the image registration algorithm proposed by Cai~\etal~\cite{cai2019toward}.
Since originally designed for image pairs taken at different focal lengths, it is hard to register stereo pairs where optical axis are not coincident.
This problem is further compounded by the unique degradation of UDC images.
As shown in Figure~\ref{fig:stitch_pairs},  Cai~\etal cannot achieve accurate registration, while our AlignFormer perfectly aligns the reference images.

\section{Additional Visual Results}
\begin{figure*}[t]
    \centering
    \includegraphics[width=.98\linewidth]{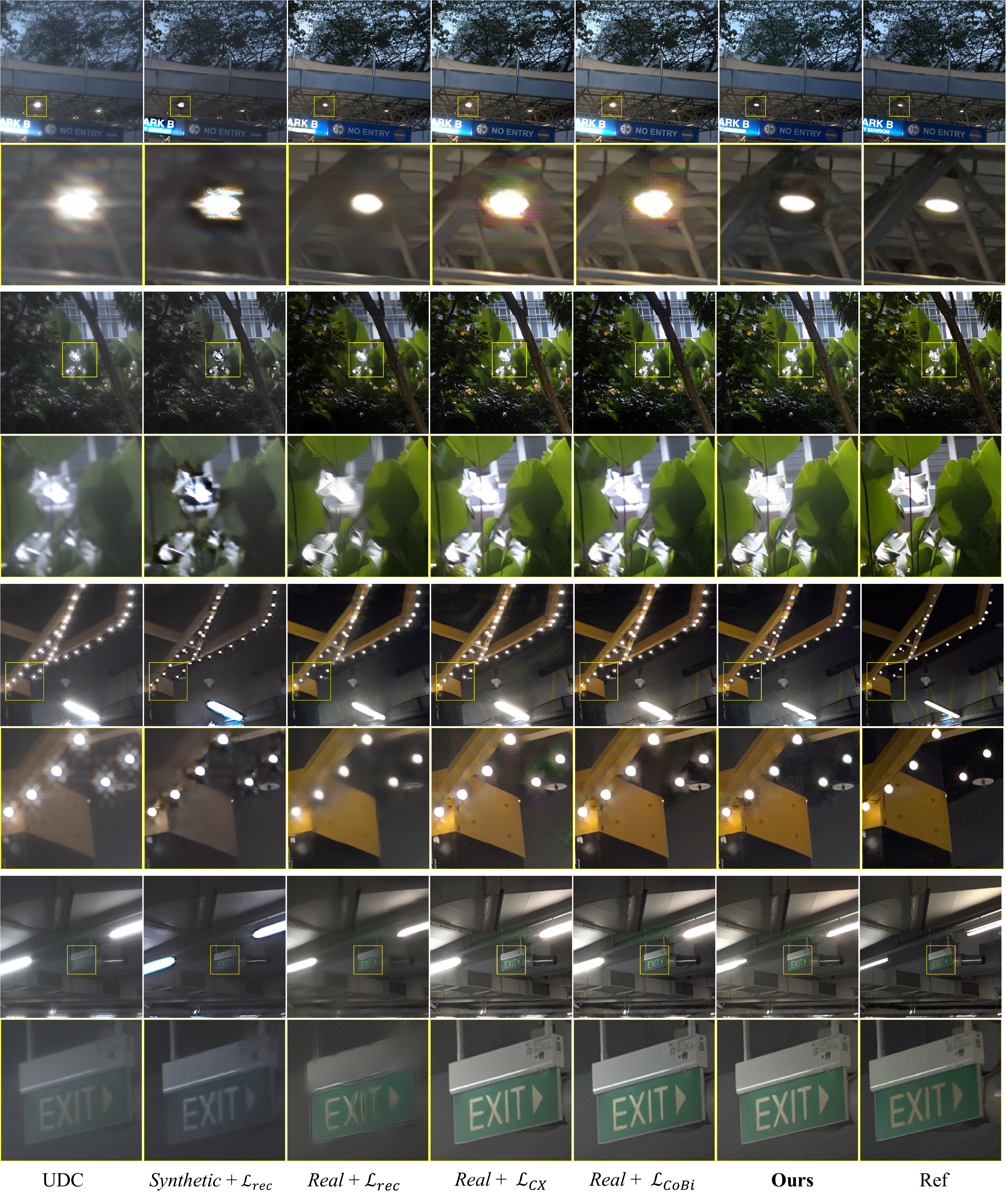}
    \vspace{-0.3cm}
    \caption{\textbf{Visual comparison between different datasets on the baseline network.}
    }
    \label{fig:add_dataset_comparison_1}
\end{figure*}

\begin{figure*}[t]
    \centering
    \includegraphics[width=.98\linewidth]{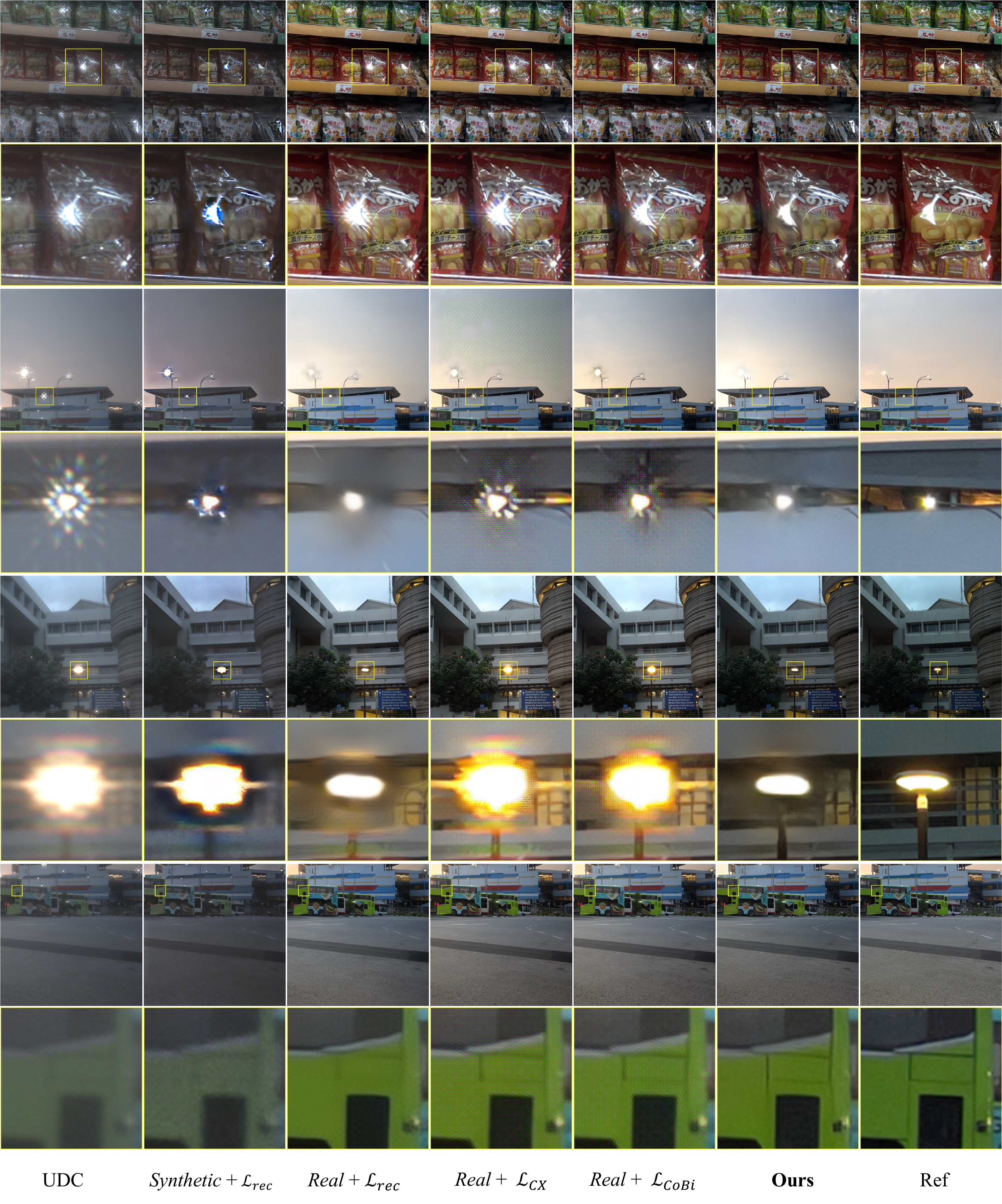}
    \vspace{-0.3cm}
    \caption{\textbf{Visual comparison between different datasets on the baseline network.}
    }
    \label{fig:add_dataset_comparison_2}
\end{figure*}

\begin{figure*}[t]
    \centering
     \includegraphics[width=0.8\linewidth]{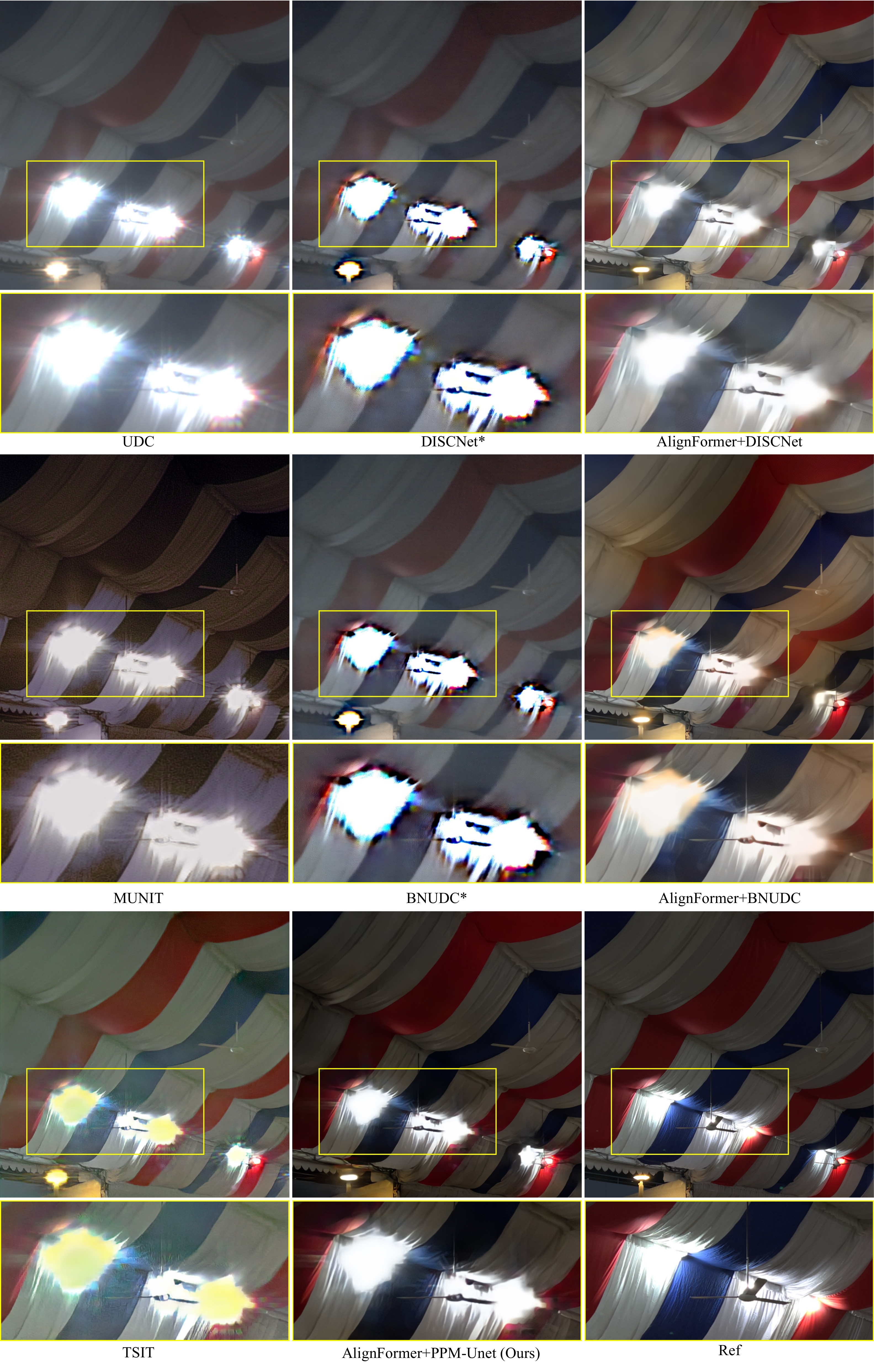}
     \vspace{-0.3cm}
    \caption{Qualitative comparisons on representative real-world samples.}
    \label{fig:add_sota_2}
\end{figure*}

\begin{figure*}[t]
    \centering
     \includegraphics[width=0.75\linewidth]{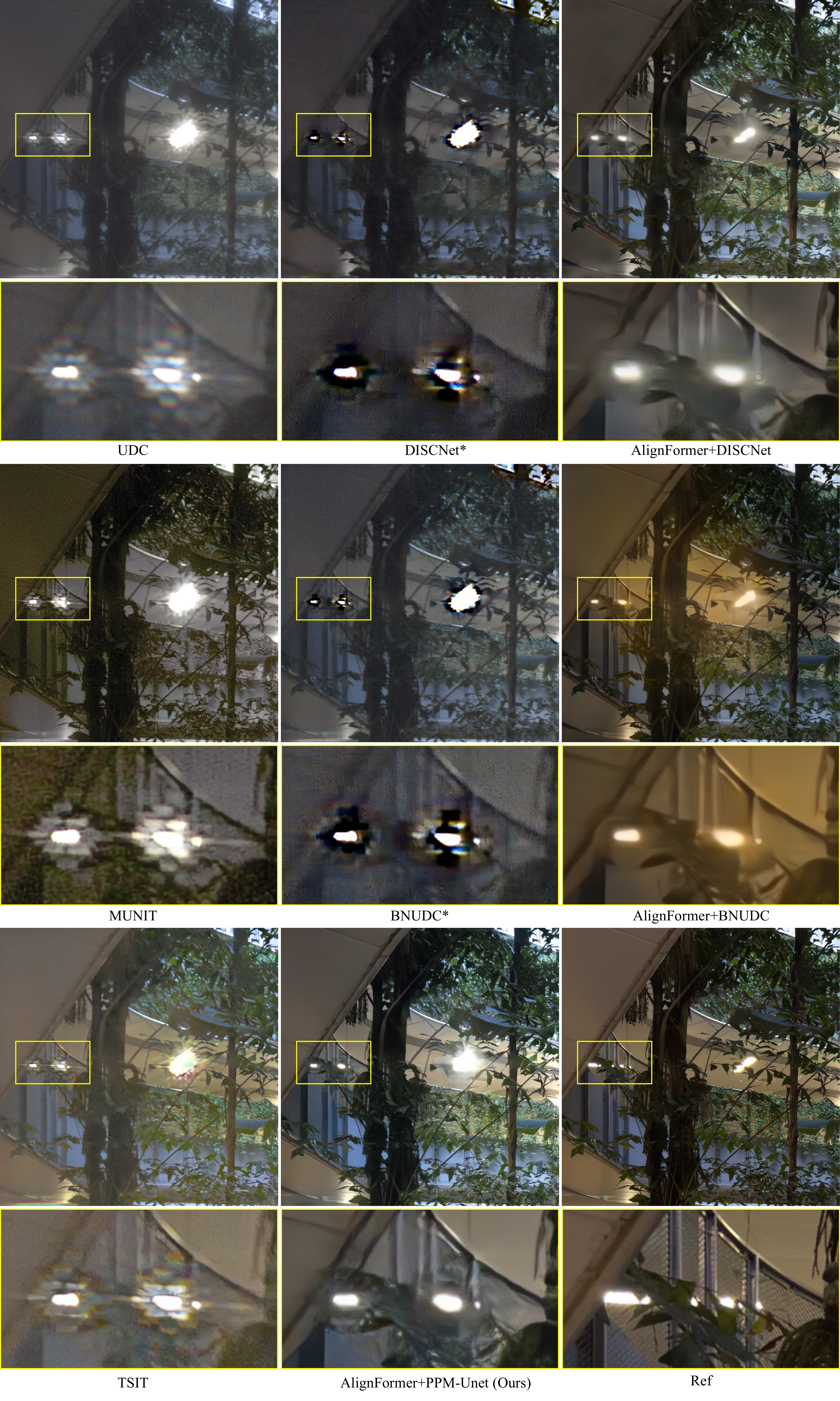}
     \vspace{-0.3cm}
    \caption{Qualitative comparisons on representative real-world samples.}
    \label{fig:add_sota_1}
\end{figure*}

We provide more visual comparisons on real data in Figure~\ref{fig:add_dataset_comparison_1} and Figure~\ref{fig:add_dataset_comparison_2}.
In addition, Figure~\ref{fig:add_sota_2} and \ref{fig:add_sota_1} present more visual comparisons with representative works.
Our method outperforms previous approaches in both removing artifacts and suppressing flare. Other methods fail to remove complicated artifacts or introduce over-correct artifacts), or produce blurry results.
The visual results suggest that our proposed data generation framework could facilitate diffraction removal and restore texture details well.

\section{Limitations}
Although the PPM-UNet, as a baseline model, already achieves promising performance for restoring real UDC images, more domain-specific designs such as aiming at the limited dynamic range of UDC images and remedying the loss of details around the over-saturation regions are required for better restoration.
While achieving rather satisfactory results on small areas of light sources, our work still struggles when highlight regions are large and intensities are extremely strong, leading to blurry results.
%
This requires further exploration on extreme cases with large and strong highlights.


\section{Broader Impacts}
Our work can be used to generate other types of aligned pseudo-supervision from non-aligned data \eg low-light/normal-light data and low-resolution/high-resolution data.
Our data will enable neural networks for restoring UDC images.
Our method will provide a new solution to such issues for 
both academia and industry.
While collecting our dataset, we try to avoid people to ensure privacy.
Therefore, our dataset does not involve ethical issues.
Moreover, as a typical image restoration task, our work will not bring negative impacts to the society.

{\small
\bibliographystyle{ieee_fullname}
\bibliography{egbib}
}


